\documentclass[10pt,twocolumn,letterpaper]{article}

\usepackage{cvpr}              %

\usepackage[dvipsnames]{xcolor}

\definecolor{cvprblue}{rgb}{0.21,0.49,0.74}
\usepackage[pagebackref,breaklinks,colorlinks,citecolor=cvprblue]{hyperref}

\def\paperID{12187} %
\def\confName{CVPR}
\def\confYear{2024}

\usepackage{float}

\usepackage{booktabs}
\usepackage{listings}
\usepackage{subcaption}
\usepackage{tabularx}
\usepackage[accsupp]{axessibility}

\usepackage{adjustbox} %
\usepackage{pifont}
\usepackage{paralist}

\newcommand{\cmark}{{\color{green} \ding{51}}} %
\newcommand{\xmark}{{\color{red} \ding{55}}} %

\usepackage{float}

\usepackage{booktabs}
\usepackage{listings}
\usepackage{subcaption}

\usepackage{algorithm}
\usepackage{algpseudocode}
\usepackage{tabularx}
\usepackage{adjustbox} %
\usepackage{pifont}
\usepackage{multicol}
\usepackage{mfirstuc}
\usepackage{float}

\usepackage{xparse}
\newcommand{\highlight}[2]{\colorbox[HTML]{#1}{#2}}

\NewDocumentCommand{\PredictionFigure}{mm}{%
\begin{figure*}[t]
  \centering
  \begin{subfigure}[c]{0.35\textwidth}
    \includegraphics[width=\linewidth, height=3cm, keepaspectratio]{images/supplementary/predictions/#11_cropped.pdf}
  \end{subfigure}%
  \begin{subfigure}[c]{0.20\textwidth}
    \includegraphics[width=\linewidth, height=3cm, keepaspectratio]{images/supplementary/predictions/#11.png}
  \end{subfigure}%
  \begin{subfigure}[c]{0.35\textwidth}
    \includegraphics[width=\linewidth, height=3cm, keepaspectratio]{images/supplementary/predictions/#12_cropped.pdf}
  \end{subfigure}%
  \begin{subfigure}[c]{0.20\textwidth}
    \includegraphics[width=\linewidth, height=3cm, keepaspectratio]{images/supplementary/predictions/#12.png}
  \end{subfigure}
  
  \vspace{0.02\textwidth} 
  
  \begin{subfigure}[c]{0.35\textwidth}
    \includegraphics[width=\linewidth, height=3cm, keepaspectratio]{images/supplementary/predictions/#13_cropped.pdf}
  \end{subfigure}%
  \begin{subfigure}[c]{0.20\textwidth}
    \includegraphics[width=\linewidth, height=3cm, keepaspectratio]{images/supplementary/predictions/#13.png}
  \end{subfigure}%
  \begin{subfigure}[c]{0.35\textwidth}
    \includegraphics[width=\linewidth, height=3cm, keepaspectratio]{images/supplementary/predictions/#14_cropped.pdf}
  \end{subfigure}%
  \begin{subfigure}[c]{0.20\textwidth}
    \includegraphics[width=\linewidth, height=3cm, keepaspectratio]{images/supplementary/predictions/#14.png}
  \end{subfigure}
  
  \caption{
    More samples from the \textsf{%
    \capitalisewords{#1}%
    } %
    benchmark.
    Legend:
    \highlight{5078e1}{OWL (L/14)},
    \highlight{2db5cb}{OWLv2 (L/14)},
    \highlight{6ee5c0}{Detic},
    \highlight{aeedaa}{ViLD},
    \highlight{ddc17f}{GDino},
    \highlight{e67951}{Cora}
}
\label{fig:additional-samples:#1}
\end{figure*}
}

\lstdefinestyle{mystyle}{
  language=Python,
  basicstyle=\ttfamily \small,
  keywordstyle=\color{blue},
  commentstyle=\color{green},
  numbers=left,
  numberstyle=\tiny\color{gray},
  breaklines=true,
  showstringspaces=false,
  frame=single
}
\title{The devil is in the fine-grained details:\\ Evaluating open-vocabulary object detectors for fine-grained understanding}

\author{
\textbf{Lorenzo Bianchi$^{1,2}$, Fabio Carrara$^1$, Nicola Messina$^1$, Claudio Gennaro$^1$, Fabrizio Falchi$^1$} \\
$^1$CNR-ISTI, Pisa, Italy \quad $^2$University of Pisa, Italy\\
\tt\small <name.surname>@isti.cnr.it
}%
\begin{document}%
\twocolumn[{%
\renewcommand\twocolumn[1][]{#1}%
\maketitle
\begin{center}
    \centering
    \captionsetup{type=figure}
    \newcommand{\teaserHeight}{5.9cm}
    \includegraphics[height=\teaserHeight]{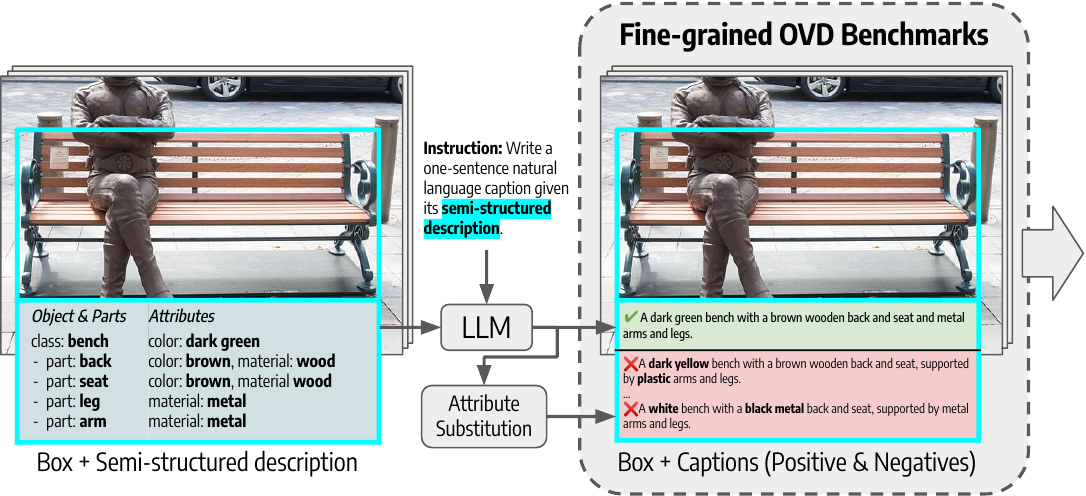} \hfill
    \includegraphics[height=\teaserHeight]{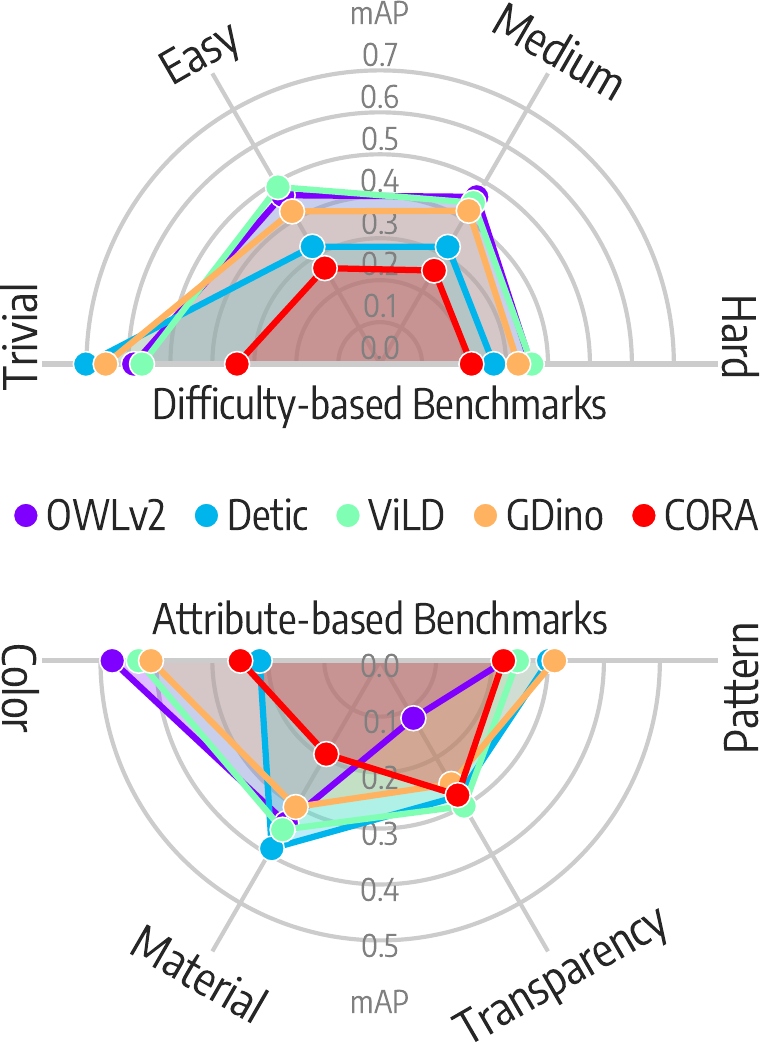}
    \captionof{figure}{
    We propose a benchmark suite to evaluate \textbf{fine-grained open-vocabulary detection (FG-OVD)}. %
    We build several sets of dynamic object-specific vocabularies, comprised of one positive and several negative captions, to probe the ability of open-vocabulary detectors to discern detailed object properties, like color, pattern, or material. %
    We craft positive captions from semi-structured descriptions of objects and their parts employing a Large Language Model (\textbf{LLM}), while negative captions of different difficulty levels are built via attribute substitution. %
    By manipulating negative sets according to their difficulty levels or the types of attributes altered --- categorized as \textbf{Difficulty-based} and \textbf{Attribute-based} benchmarks --- we acquire a nuanced comprehension of each detector's capabilities across various scenarios.%
    }
    \label{fig:teaser}
\end{center}%
}]

\begin{abstract}

Recent advancements in large vision-language models enabled visual object detection in open-vocabulary scenarios, where object classes are defined in free-text formats during inference.
In this paper, we aim to probe the state-of-the-art methods for open-vocabulary object detection to determine to what extent they understand fine-grained properties of objects and their parts.
To this end, we introduce an evaluation protocol based on dynamic vocabulary generation to test whether models detect, discern, and assign the correct fine-grained description to objects in the presence of hard-negative classes.
We contribute with a benchmark suite of increasing difficulty and probing different properties like color, pattern, and material.
We further enhance our investigation by evaluating several state-of-the-art open-vocabulary object detectors using the proposed protocol and find that most existing solutions, which shine in standard open-vocabulary benchmarks, struggle to accurately capture and distinguish finer object details.
We conclude the paper by highlighting the limitations of current methodologies and exploring promising research directions to overcome the discovered drawbacks. Data and code are available at \href{https://lorebianchi98.github.io/FG-OVD/}{https://lorebianchi98.github.io/FG-OVD/}
.

\end{abstract}

\section{Introduction}

    Open-vocabulary object detection (OVD) consists of recognizing objects not present at training time, therefore solving the limitations imposed by traditional detectors that could only recognize a fixed pool of object classes.
    In the last years, open-vocabulary detectors have captured large research attention thanks to their wide flexibility in many potential downstream applications like autonomous driving \cite{ma2022rethinking}, extended reality \cite{lu2023open}, and robotics \cite{liang2023code, huang2023voxposer, jatavallabhula2023conceptfusion}.
    The core idea behind open-vocabulary detectors is to establish a semantic connection between object regions and object labels chosen from a possibly very large vocabulary by (i) employing some focused prompting on the object label to transform it into a natural sentence and (ii) relying on vision-language interaction --- possibly employing large pre-trained matching methods like CLIP \cite{radford2021learning} --- for associating the most relevant object label with the specific image region.

    Given the vision-language nature of open-vocabulary detectors, the elements in the labels' vocabulary are no longer restricted to simple categorical values.
    Instead, labels can be seamlessly substituted with any natural language sentence, ranging from compound objects possibly containing modifying attributes --- i.e., for distinguishing a \textit{brown bear} from a \textit{polar bear} --- to very complex sentences describing extrinsic object properties like colors, patterns, or materials.
    Therefore, we may ideally pretend that a pre-trained open-vocabulary detector is also able --- to some extent --- to discern a \textit{dark brown wooden lamp} from a \textit{gray metal lamp}. This is what we call \textbf{F}ine-\textbf{G}rained\footnote{Rather than closed-set fine-grained categorization~\cite{wei2021fine} focusing on small inter-class variation (e.g., \textit{golden retriever} vs \textit{labrador retriever}), we focus on intra-class extrinsic attributes (e.g., \textit{brown dog} vs \textit{black dog}).} \textbf{O}pen-\textbf{V}ocabulary object \textbf{D}etection (\textbf{FG-OVD}).
    
    Apart from some marginal attempts \cite{ramanathan2023paco,bravo2023open}, no work has deeply investigated the ability of open-vocabulary detectors to discern fine-grained object properties.
    Some emerging downstream tasks start to require this ability, like episodic memory tasks in egocentric videos\footnote{\url{https://github.com/EGO4D/episodic-memory}}, where it is asked to retrieve object \textit{instances} having particular attributes.
    
    However, it is difficult to evaluate the ability of an open-vocabulary detector to discern specific object properties. %
    As of now, most works evaluate the effectiveness of open-vocabulary detectors using established benchmarks like COCO and LVIS, which are already widely used to train and test traditional object detectors.
    Given their closed-vocabulary nature, these benchmarks %
    primarily focus on very generic class labels and do not explore the capabilities of these detectors when the input text is more elaborate and includes fine-grained characteristics of the object. %

    In this paper, we propose a novel evaluation protocol and %
    a benchmark suite for measuring the discriminative power of open-vocabulary detectors against fine-grained object descriptions.
    We build a set of object detection benchmarks that provide each object with a rich and intricate caption --- generated by a Large Language Model (LLM) fed with a structured description of objects in the scene --- that encapsulates its complex extrinsic characteristics like color, material, or pattern.
    For each positive caption, we also include a series of incorrect variants obtained through slight attribute modifications, acting as negative labels. %

    By carefully varying the number and types of attributes in the negative examples for each object, our benchmarks, together with the proposed evaluation protocol, enable us to analyze the weaknesses of the most recent open-vocabulary detectors across different analytical dimensions, identifying areas for improvement and providing valuable directions for advancing this extremely relevant field.

    In summary, our key contributions are as follows:
    \begin{compactitem}
        \item We introduce the novel challenging task of Fine-grained Open Vocabulary Detection (FG-OVD).
        \item We propose a novel evaluation protocol for FG-OVD, complimented by some relevant metrics, for quantitatively assessing the fine-grained discriminative power of open-vocabulary detectors.
        \item We introduce a novel set of benchmarks specifically crafted to evaluate the ability of open-vocabulary detectors to discern extrinsic object properties.
        \item We perform extensive experimentation, testing some state-of-the-art pre-trained open-vocabulary detectors and demonstrating that even the most recent ones struggle to distinguish fine-grained object properties.
    \end{compactitem}

\section{Related work}
    
    \subsection{Zero-Shot Open-vocabulary Object detection}
        Zero-shot (ZS) generically refers to scenarios where the model, at inference time, is not exposed to specific object classes seen during its training process. In the context of object detection, we follow the one given by~\cite{arandjelovic2023three}, where ZS refers to never seeing, during training, even a single annotated bounding box of the class of interest at test time. \footnote{In our case, we deal with free-form sentences as labels and not classical categorical ones, which makes all the possible phrasings of every combination of attributes almost unique. Therefore, we deem the proposed task lies in the ZS regime since we have as labels, in the test set, a set of sentences comprising attributes unlikely seen during training.} %
        
        Early approaches in zero-shot object detection, such as Bansal et al.~\cite{bansal2018zero}, proposed replacing the last classification layer with language embeddings, such as GloVe~\cite{pennington2014glove}, representing class names.
        However, recent advancements in large-scale image-text encoder models --- such as CLIP~\cite{radford2021learning} and ALIGN~\cite{jia2021scaling}, trained on millions of image-text pairs --- enabled a strong semantic interaction between vision and language.
        Their cross-modal alignment capabilities have been used in a plethora of recent open-vocabulary detectors by either substituting the final class features with their text embeddings~\cite{zhou2022detecting,minderer2022simple,feng2022promptdet,kuo2022f, minderer2024scaling}, by learning better ROI-Align head~\cite{he2017mask} using knowledge from pre-trained vision-language backbones~\cite{gu2021open, du2022learning, zhou2022learning, kuo2022f, minderer2024scaling}, or by directly repurposing the CLIP model itself to work as an open-vocabulary detector~\cite{zhong2022regionclip,minderer2022simple,minderer2024scaling}. 

        Some recent works \cite{li2022grounded,liu2023grounding,kamath2021mdetr,zhang2022glipv2} started to unify open-vocabulary detection with the Referring Expression Comprehension (REC)~\cite{yu2016modeling,mao2016generation} and Phrase Grounding (PG)~\cite{mu2021disentangled,plummer2015flickr30k} tasks, already well-known in literature.
        While both REC and PG are given a single possibly complex sentence, REC's objective is to locate the single correct object in the image, while PG requires locating all the entities appearing in the text.
        Despite the similarities with the FG-OVD task and their higher need for associating articulated phrases to image regions, there are some key differences: (i) REC and PG tasks assume that the given unambiguous sentence is surely grounded somewhere in the image, making the development of advanced discriminative abilities almost unnecessary; (ii) networks trained to solve these tasks are not evaluated on their discriminative abilities when posed with difficult choices.
        Therefore, although some REC or PG networks can also work as open-vocabulary detectors --- like the recent GroundingDino \cite{liu2023grounding} or GLIP \cite{zhang2022glipv2,li2022grounded} --- they tend to inherit the same drawbacks when they are required to distinguish fine-grained object characteristics.

    \subsection{Open-Vocabulary Detection Benchmarks}
        
        The COCO~\cite{lin2014microsoft} and LVIS~\cite{gupta2019lvis} datasets are widely recognized as benchmarks for evaluating the localization and classification capabilities of object detectors.
        The COCO dataset, originally largely employed for evaluating standard closed-set detectors, has been repurposed to address zero-shot detection~\cite{bansal2018zero} and open-vocabulary detection~\cite{gu2021open,zhong2022regionclip,liu2023grounding}.
        In this setup, it is composed of a vocabulary of 48 base categories for training and 17 novel categories for testing.
        On the other hand, LVIS contains a large and diverse collection of object categories grouped by frequency of appearance (common, frequent, and rare).
        Many works~\cite{gu2021open, zhong2022regionclip, kuo2022f, liu2023grounding} used the frequent and common categories as base categories and rare categories as novel categories for testing.
        While these benchmarks evaluate the model's ability to perform zero-shot open-vocabulary detection, they do not explicitly evaluate the model's ability to recognize specific object characteristics.

        The datasets that most closely align with our objectives are OVAD~\cite{bravo2023open} and VAW~\cite{pham2021learning}.
        Despite focusing on object attributes and proposing to use negatives to challenge existing detectors, there are major differences with our work:
        (i) they mainly benchmark attribute detectors, which usually have a separate head to specifically infer attributes besides the object classes;
        (ii) they only need structured annotations and not natural language sentences describing each object, limiting the evaluation of current state-of-the-art vision-language foundation models;
        (iii) they do not craft challenging negative examples, limiting the analysis of the current detector's limitations.
        The dataset from which we drew inspiration is PACO~\cite{ramanathan2023paco}.
        PACO is built upon COCO and comes with object bounding boxes annotated with a structured JSON-like representation that carries information about attributes and parts information of the object.
        PACO carries extrinsic and categorized object properties%
        , like object colors, materials, and patterns.
        Our benchmarks suite is crafted from PACO, by creating fine-grained sentences from structured object descriptions using LLMs, empowering the growing trend of using LLMs to create targeted, high-quality, and diverse annotations \cite{momeni2023verbs}. %

\section{Methodology}

    \begin{figure}
        \centering
        \includegraphics[width=0.9\linewidth]{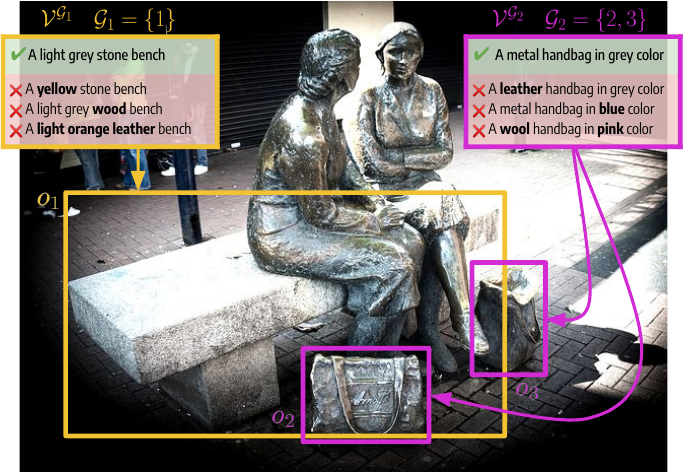}
        \caption{\textbf{Examples of Dynamic Vocabularies:} the image $I$ features two distinct object groups $\mathcal{G}_1$ and $\mathcal{G}_2$, each one associated with a set of captions. The positive captions $c^\text{pos}$ (marked with \cmark %
        ) -- \textit{A light grey stone bench} and \textit{A metal handbag in grey color} --, are juxtaposed with three negative captions $c^\text{neg}$ (indicated by \xmark %
        ). These positive and negative captions collectively form two vocabularies, namely $\mathcal{V}^{\mathcal{G}_1}$ assigned to $o_1$, and $\mathcal{V}^{\mathcal{G}_2}$ assigned to $o_2$ and $o_3$. The open-vocabulary detector is then applied to $I$ two times, once for each vocabulary: $\psi(I, \mathcal{V}^{\mathcal{G}_1})$ and $\psi(I, \mathcal{V}^{\mathcal{G}_2})$.}
        \label{fig:pacco-example}
    \end{figure}

    We propose an evaluation protocol to probe the capabilities of open-vocabulary object detectors to discern fine-grained characteristics of objects to be detected.
    In \autoref{sec:evaluation-protocol}, we formulate the problem and propose a suitable evaluation protocol.
    Then, in \autoref{sec:dataset}, we describe the novel suite of benchmarks aimed at challenging open-vocabulary detectors to distinguish fine-grained object labels. %

    \subsection{Evaluation Protocol}
    \label{sec:evaluation-protocol}

    \paragraph{OVD Formalization.}
    \newcommand{\R}{\mathbb{R}}
    Let $I \in \R^{W \times H \times C}$ an image and $\mathcal{V} = \{c_j\}_{j=1}^{T}$ a vocabulary composed by $T$ arbitrary sentences describing the objects we want to detect.
    An open-vocabulary object detector $\psi$ takes in input $I$ and $\mathcal{V}$ and produces a set of predictions $\mathcal{D} = \psi(I,\mathcal{V}) = \{d_i\}_{i=1}^m$, where each prediction $d_i = (\mathbf{b}_i, \mathbf{s}_i)$ is comprised by a bounding box $\mathbf{b}_i \in \R^4$, and the scores $\mathbf{s}_i \in \R^T$, where each element $s_{i,j}$ is the confidence assigned to vocabulary object $c_j$
    \footnote{Usually, captions are obtained by prompting a vocabulary of class labels (e.g., \textit{dog} $\to$ \textit{a photo of a dog}).}%
    .
    The highest-score caption in $\mathcal{V}$ is selected as the predicted one.  %
    In classic zero-shot and open-vocabulary evaluation settings, the vocabulary $\mathcal{V}$ is fixed for all images in the test dataset, and for each image, predictions $\mathcal{D}$ are compared with the ground truth object instances $\mathcal{O} = \{o_i\}_{i=1}^{m'},\,o_i = (\mathbf{b'}_i, l'_i) \in \R^4 \times \mathcal{V}$ and evaluated by means of standard detection metrics, e.g., mAP, mAR, etc.

    \paragraph{Dynamic Vocabularies for FG-OVD.}
    For evaluating FG-OVD we suggest utilizing a carefully crafted dictionary $\mathcal{V}_i$ for each ground truth object $o_i$. This approach allows us to have finer control over the selection of negatives to be used for each object. %
    More formally, each object $o_i$ is given a detailed visual description denoted as $c^\text{pos}_i$ (referred to as the positive caption) that accurately describes an object.
    The object $o_i$ is also associated with several other descriptions $c^\text{neg}_{i,1}, \dots, c^\text{neg}_{i,N}$ (referred to as negative captions) that semantically differ --- up to varying degrees --- from $c^\text{pos}_i$. In this way, we define a new ground truth composed of objects $\tilde{o}_i = (\mathbf{b'}_i, l'_i, \mathcal{V}_i)$, where $\mathcal{V}_i$ is a per-object vocabulary filled with $\{c^\text{pos}_i, c^\text{neg}_{i,1}, \dots, c^\text{neg}_{i,N}\}$, and the label $l'_i = c^\text{pos}_i$.
    The detector draws the predicted label from the vocabulary $\mathcal{V}_i$ during inference.
    From a practical perspective, notice that if we perform inference for each object $o_i$ over its vocabulary $\mathcal{V}_i$ inside the same image $I$, we may produce duplicated outputs in the case where there are two objects $o_i$ and $o_j$ both correctly described by the same $c^\text{pos}$. In order to solve this problem, we perform a single inference for each one of the $K$ set of objects $o_{\mathcal{G}_k}$ in the image $I$ sharing the same positive caption $c^\text{pos}$, where $\mathcal{G}_k$ is the $k$-th set of object indexes satisfying this property. Therefore, we perform $K$ separate inferences for each set $\mathcal{G}_k$ by performing the forward passes $\{\psi(I, \mathcal{V}^{\mathcal{G}_k})\}_{k=1}^K$, where $\mathcal{V}^{\mathcal{G}_k}$ is the vocabulary associated with the objects $o_{\mathcal{G}_k}$.
    We report an example of our ground truth arrangement in \autoref{fig:pacco-example}.

    \paragraph{Post-processing.} %
    Class-aware non-maximum suppression (NMS) is typically employed in object detection post-processing to discard near-duplicate predictions of the same class insisting on the same location, while different-class predictions of the same object are left untouched.
    In our formulation, classes in the vocabulary are mutually exclusive, and we are interested in evaluating only the most confident prediction per object independently from the class label.
    Traditional implementations of detection metrics, such as COCO mAP, do not penalize incorrect predictions of non-occurring classes, such as the negative ones in our vocabularies, i.e., the presence of an incorrect higher-confidence prediction in the same location as a correct prediction is not penalized.
    To circumvent this issue, we apply \emph{class-agnostic} NMS to ensure only one prediction per location is present, regardless of class label.
    This guarantees that an incorrect higher-confidence prediction suppresses the lower-confidence correct prediction, correctly bringing the mAP metric to work as desired in our evaluation setup.

    \paragraph{Metrics.}
    The post-processed predictions are rigorously evaluated using the COCO mean Average Precision (mAP) metric, which provides a comprehensive assessment of object localization accuracy and caption assignment correctness across varying detection confidence levels. 
    
    Alongside mAP, we incorporate the Median Rank metric.
    Given an object $o_i = (\mathbf{b'}_i, l'_i)$ and its corresponding vocabulary $\mathcal{V}_i$, let $d^{i}_j = (\mathbf{b}_j, \mathbf{s}_j)$ the prediction %
    satisfying the condition of Intersection over Union (IoU) $\geq 0.5$ with $o_i$ \footnote{In rare edge cases, we might have multiple predictions insisting on the same ground truth box not removed by NMS. In those cases, we select the prediction with the maximum highest confidence.}.
    We sort the confidence scores of each element of the vocabulary $\mathbf{s}_j$ in descending order and record the position of the score corresponding to the correct caption $l'_i$ in the ranked list.
    Then, we report the median rank over all objects in the benchmark.
    
    Unlike mAP, which only considers the maximally activated label, the median rank helps to better quantify the confidence of each detector in predicting the correct label among the other choices available in the dictionary.

\begin{table*}[t]
    \centering
    \caption{Statistics of the benchmarks for each different negative set comprising the number of images (Imgs), the number of annotated objects (Objs), objects-to-image ratio (Objs/Img), positive captions, positive captions per image, negative captions per positive caption, and objects per positive caption.}
    \label{tab:benchmark_statistics}
    \begin{adjustbox}{center}
    \begin{tabular}{llrrcrcrc}
    \toprule
    Name         & Negative Set Strategy                &  Imgs &    Objs &    Obj/Img & \cmark Caps & \cmark/Img & \xmark/\cmark &   Objs/\cmark \\
    \midrule
    Hard         & Random attribute subst. $(\times 1)$ &  1707 &    3545 &        2.1 &        2349 &        1.4 &           9.9 &           1.5 \\
    Normal       & Random attribute subst. $(\times 2)$ &  1537 &    2968 &        1.9 &        2034 &        1.3 &          10.0 &           1.5 \\
    Easy         & Random attribute subst. $(\times 3)$ &   853 &    1299 &        1.5 &         971 &        1.1 &          10.0 &           1.3 \\
    Trivial      & Random captions                      &  1707 &    3545 &        2.1 &        2349 &        1.4 &           9.9 &           1.5 \\
    \midrule
    Color        & Color attribute subst.               &  1599 &    3119 &        2.0 &        2126 &        1.3 &          10.0 &           1.5 \\
    Material     & Material attribute subst.            &  1577 &    3193 &        2.0 &        2128 &        1.3 &          10.0 &           1.5 \\
    Pattern      & Pattern attribute subst.             &   321 &     467 &        1.5 &         337 &        1.0 &           7.4 &           1.4 \\
    Transparency & Transparency attribute subst.        &   230 &     409 &        1.8 &         238 &        1.0 &           2.2 &           1.7 \\
    \bottomrule
    \end{tabular}
    \end{adjustbox}
    \end{table*}

    \begin{figure*}[t]
        \captionsetup{justification=centering}
        \newcommand{\colwid}{0.225\linewidth}
        \centering
        \sffamily
        \begin{subfigure}[t]{\colwid}
            \caption{\textsf{Trivial}} %
            \label{fig:examples:trivial}
            \includegraphics[width=\linewidth]{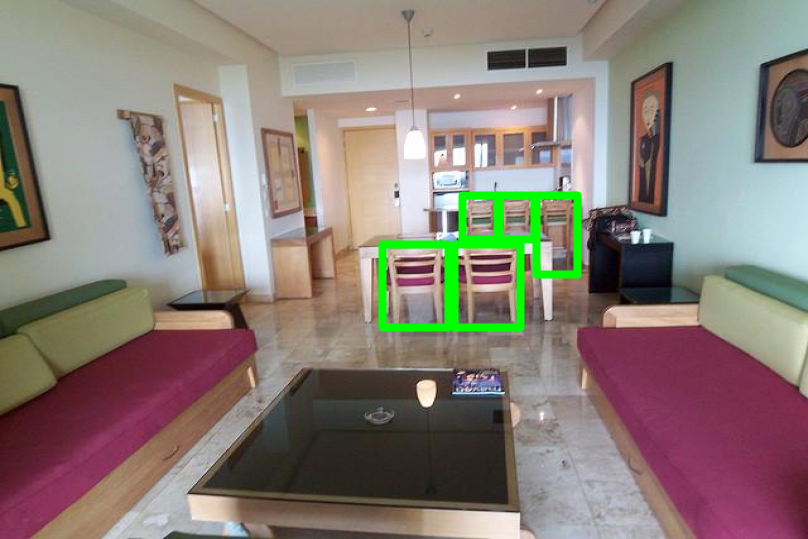}
            \tiny\raggedright
            \cmark A brown wooden chair.\\
            \xmark\ \underline{A red cup with a pink plastic rim}\\
            \xmark\ \underline{A pink dog with a black and dotted ear.} \\
        \end{subfigure}
        \hfill
        \begin{subfigure}[t]{\colwid}
            \caption{\textsf{Easy}}%
            \label{fig:examples:easy}
            \includegraphics[width=\linewidth]{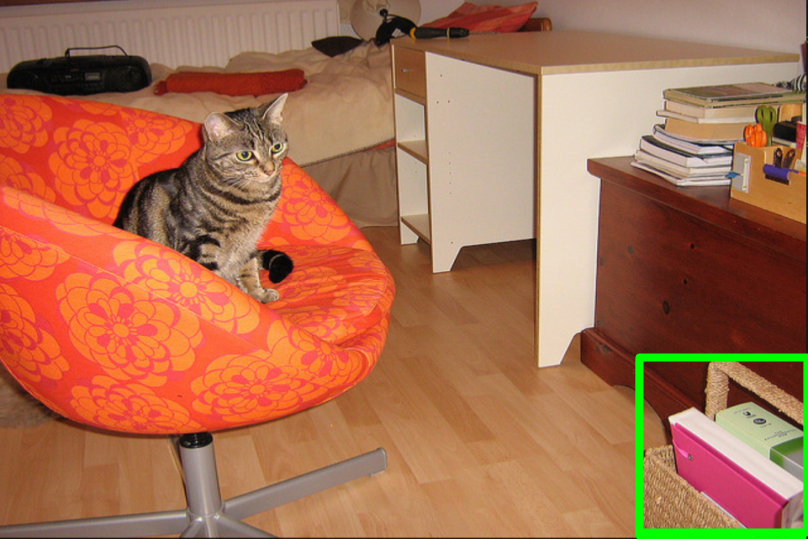}
            \tiny\raggedright
            \cmark A brown woven rattan basket.\\
            \xmark\ A \underline{light green} \underline{perforated} \underline{fabric} basket.\\
            \xmark\ A \underline{black} \underline{dotted} \underline{leather} basket. \\
        \end{subfigure}
        \hfill
        \begin{subfigure}[t]{\colwid}
            \caption{\textsf{Medium}}%
            \label{fig:examples:medium}
            \includegraphics[width=\linewidth]{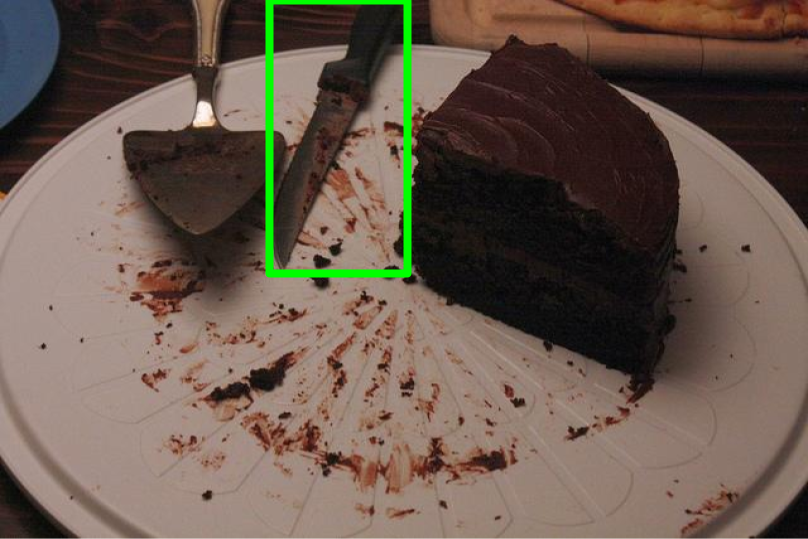}
            \tiny\raggedright
            \cmark A knife with a black plastic handle and a dark grey metal blade.\\
            \xmark\ A knife with a \underline{grey} \underline{stone} handle and a dark grey metal blade.\\
            \xmark\ A knife with a \underline{light pink} plastic handle and a \underline{light yellow} metal blade.\\
        \end{subfigure}
        \hfill
        \begin{subfigure}[t]{\colwid}
            \caption{\textsf{Hard}}%
            \label{fig:examples:hard}
            \includegraphics[width=\linewidth]{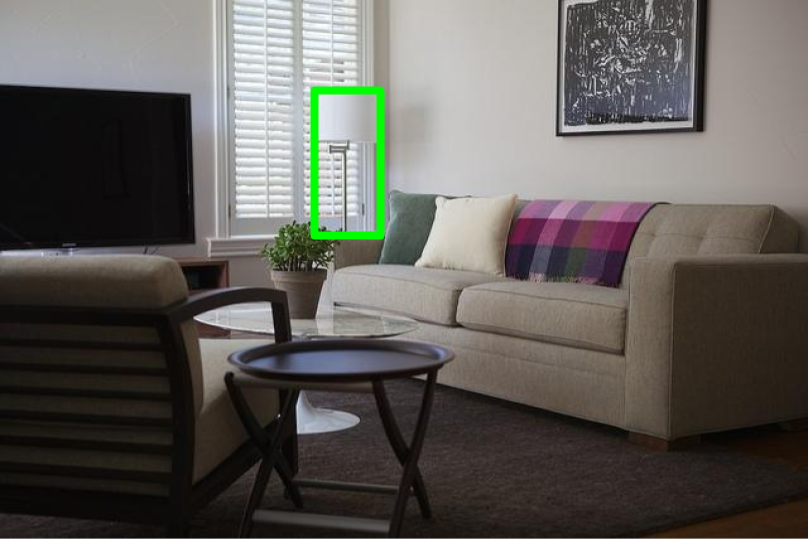}
            \tiny\raggedright
            \cmark A lamp with a white plastic shade and a grey metal pipe.\\
            \xmark\ A lamp with a white \underline{velvet} shade and a grey metal pipe.\\
            \xmark\ A lamp with a white plastic shade and a \underline{dark pink} metal pipe. \\
        \end{subfigure}
        \begin{subfigure}[t]{\colwid}
            \caption{\textsf{Color}}%
            \label{fig:examples:color}
            \includegraphics[width=\linewidth]{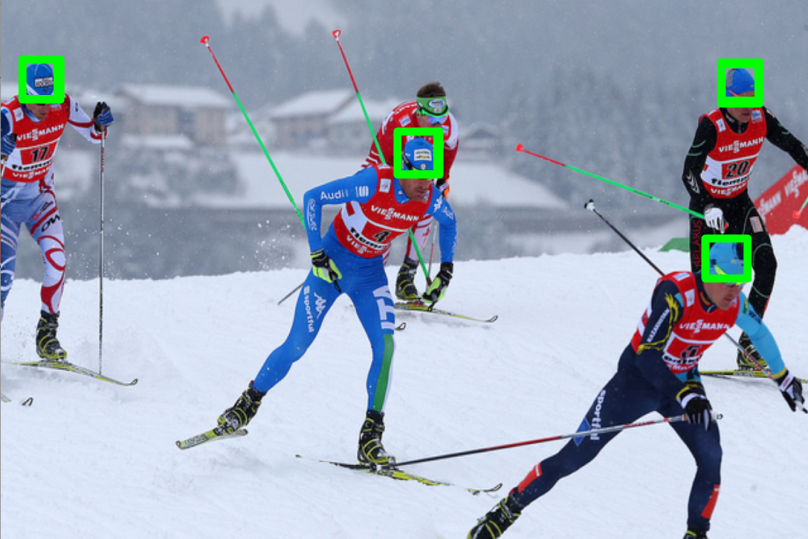}
            \tiny
            \cmark A blue hat. \\
            \xmark\ A \underline{orange} hat. \\
            \xmark\ A \underline{yellow} hat. \\
        \end{subfigure}
        \hfill
        \begin{subfigure}[t]{\colwid}
            \caption{\textsf{Material}}%
            \label{fig:examples:material}
            \includegraphics[width=\linewidth]{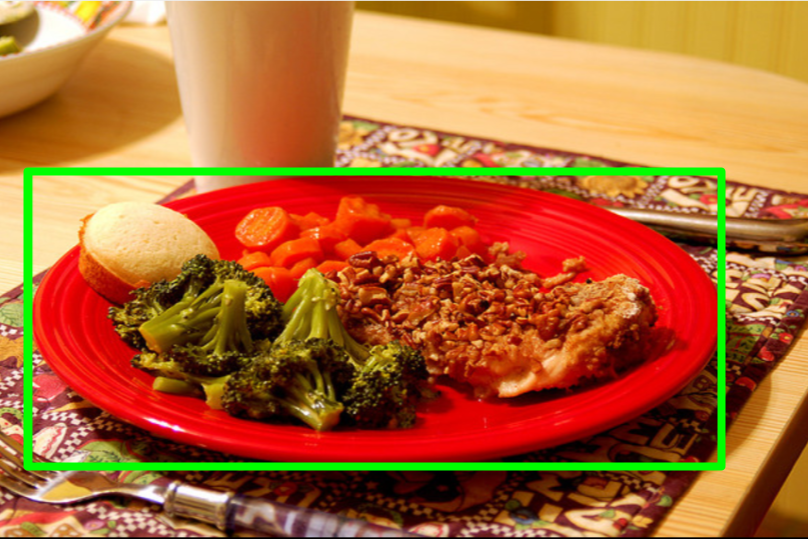}
            \tiny
            \cmark A red plastic plate.\\
            \xmark\ A red \underline{metal} plate.\\
            \xmark\ A red \underline{ceramic} plate.\\
        \end{subfigure}
        \hfill
        \begin{subfigure}[t]{\colwid}
            \caption{\textsf{Pattern}}%
            \label{fig:examples:pattern}
            \includegraphics[width=\linewidth]{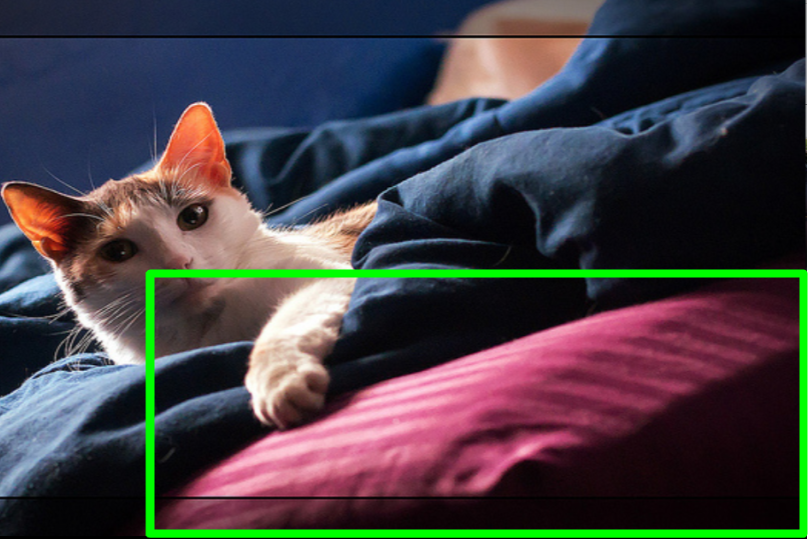}
            \tiny\raggedright
            \cmark A dark pink striped pillow made of fabric.\\
            \xmark\ A dark pink \underline{floral} pillow made of fabric.\\
            \xmark\ A dark pink \underline{dotted} pillow made of fabric.\\
        \end{subfigure}
        \hfill
        \begin{subfigure}[t]{\colwid}
            \caption{\textsf{Transparency}}%
            \label{fig:examples:transparency}
            \includegraphics[width=\linewidth]{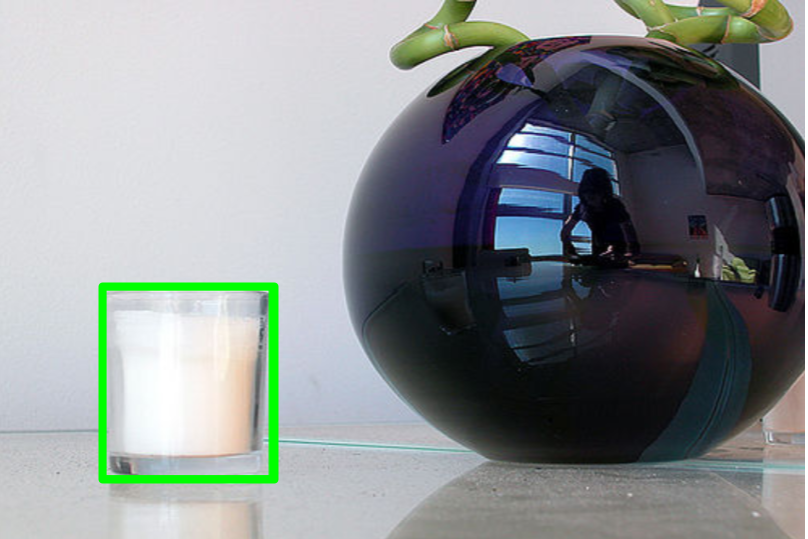}
            \tiny
            \cmark A transparent glass.\\
            \xmark\ A \underline{translucent} glass.\\
            \xmark\ A \underline{opaque} glass.\\
        \end{subfigure}\\%

        \caption{\textbf{Benchmarks Examples}: each benchmark tests different properties by crafting negative captions via attribute substitution.}
        \label{fig:examples}
    \end{figure*}

    \subsection{Dataset}
    \label{sec:dataset}
    Evaluating open-vocabulary object detectors through the aforementioned evaluation protocol requires a carefully crafted benchmark, carrying different vocabularies for each object. A benchmark implementing this novel evaluation protocol should pay particular attention to the quality and variability of negative examples in each object's vocabulary.
    To this aim, we introduce an extension to the PACO dataset \cite{ramanathan2023paco}, which comprises an innovative suite of benchmarks meticulously crafted to tackle the task of FG-OVD. Our benchmark suite offers a comprehensive evaluation through eight distinct scenarios, categorized into \textbf{Difficulty-based} and \textbf{Attribute-based} benchmarks. Difficulty-based benchmarks enable the assessment of detector performance across different difficulty levels obtained by varying the hardness of negative captions. On the other hand, Attribute-based benchmarks enable the precise selection of attribute types to facilitate the evaluation of detectors' capabilities in recognizing specific attributes. For each of the four attribute types, we mostly inherit the possible values from the PACO dataset, for a total of \textit{29 colors}, \textit{14} \textit{materials}, \textit{8} \textit{patterns}, and \textit{3} \textit{transparencies} modes (more details in the supplementary material).
    In the following paragraphs, we dive into the construction details of such benchmarks. %
    
    \paragraph{Positive Caption Generation.} 
    In our quest to describe objects with precision, we assume each object is characterized by at least one attribute and may possess one or more parts, where each part may have its own attributes. As we require text entries in open-vocabulary detectors, we exploit this structured definition of each object to generate a fine-grained textual description carrying attributes and parts details. %
    To achieve this, we harnessed the capabilities of a Large Language Model (LLM), by prompting it with structured object descriptions from PACO -- as shown in \autoref{fig:teaser} -- and forcing it to mimic some carefully crafted natural language examples. We then performed a quick manual check to ensure the required consistency.
    Notice that PACO already provides straightforward natural language captions for a limited number of objects. We still preferred the ones generated through an LLM, considering the beneficial quality-to-quantity ratio compared with the already available ones. Detailed comparisons with the original PACO captions are given in supplementary material.
    We decided to employ OpenAssistant-LLAMA-30B \cite{kopf2024openassistant} as an LLM, as it provides finer control over inference and better reproducibility guarantees over other available LLMs.

    \paragraph{Negative Captions Generation.}
    To thoroughly assess open-vocabulary object detection models, we introduce diverse benchmarks featuring vocabularies with challenging negative captions derived from the generated captions.
    We aim to create negative captions that are semantically different from the positive ones while maintaining structural similarities.
    We do so through attribute substitution, which involves replacing attributes in the positive captions text to preserve syntactic consistency.
    We preferred this approach to altering the object's structured description and re-querying the LLM, which could introduce significant syntactic variations and unwanted hallucinations.
    
    The choice of negative captions aims to assess the model's adaptability to a range of attribute variations, covering changes in attribute types and quantities.
    This allows for a comprehensive evaluation of its resilience across diverse negative scenarios.
    The benchmarks fall into two classes:
    \textbf{(i) Difficulty-based} (\textsf{Trivial, Easy, Medium, Hard}): These benchmarks explore open-vocabulary detectors' capabilities in ascending difficulty scenarios. Starting with the \textsf{Trivial} benchmark, where negative captions are randomly sampled from other objects, we progress through \textsf{Easy}, \textsf{Medium}, and \textsf{Hard} benchmarks, in which negative captions are generated by randomly replacing 3, 2, and 1 attributes, respectively. %
    As the number of attributes replaced decreases, the distinctions between captions become less pronounced and the task more challenging (see \autoref{fig:examples}, top row). \textbf{(ii) Attribute-based} (\textsf{Color, Material, Transparency, Pattern}): This class of benchmarks delves into open-vocabulary detectors' capabilities in recognizing specific types of attributes. By precisely replacing only one attribute of a specific type in each positive caption (depending on the specific category of the benchmark), we aim to evaluate the detector's proficiency in recognizing that particular attribute type (see \autoref{fig:examples}, bottom row).
    \autoref{tab:benchmark_statistics} reports statistics of each described benchmark.
    
\section{Experiments}

    \subsection{Evaluated Models}
    \label{sec:evaluated-models}
    In our experiments, we evaluated the performance of the following state-of-the-art open-vocabulary detectors.
    
    \textbf{ViLD} \cite{gu2021open}, \textbf{Detic} \cite{zhou2022detecting}, and \textbf{CORA} \cite{wu2023cora} underwent evaluation using our standard protocol outlined in \autoref{sec:evaluation-protocol} for all the images within the FG-OVD benchmarks. Specifically, for every object group $i$ in the image $I$, we perform an inference $\psi(I, \mathcal{V}^{\mathcal{G}_i})$. %
    
    \textbf{OWL} \cite{minderer2022simple} and \textbf{OWLv2} \cite{minderer2024scaling} adhered to the same standard protocol. However, due to the positional embedding layer of these models being trained with a limit of 16 tokens, evaluation was conducted exclusively on captions respecting this maximum token constraint (about 80\% of the available ones). We then checked that applying this same inference constraint to all the detectors did not consistently affect the overall evaluation. We report the details of this sanity check in the supplementary material.
    
    \textbf{GroundingDino} \cite{liu2023grounding}, is mainly a REC model and thus, in contrast to other detectors, cannot accept a vocabulary of captions as input. The authors claim that specific prompts enable the given input textual expression to be automatically split -- e.g., after each "." character. However, we found that this works only when single words are given as labels, while complex sentences are sometimes erroneously split in their middle. %
    Therefore, we turned GroundingDino into an open-vocabulary detector by making a forward pass for each of the captions in the vocabulary $\mathcal{V}_i$ and merging the results before evaluation. More details about the specific inference process are given in the supplementary material.

    \subsection{Results}
    \begin{table}
        \centering
        \setlength{\tabcolsep}{5pt}
        \newcolumntype{C}{>{\centering\arraybackslash}X}
        \begin{tabularx}{\linewidth}{l*4cC}
        \toprule
                         & \multicolumn{4}{c}{FG-OVD} & LVIS\\
                           \cmidrule(lr){2-5} \cmidrule(lr){6-6}
        Detector     &    \small Hard &  \small Medium &    \small Easy & \small Trivial & Rare \\
        \midrule
        OWL  \small (B/16)   &           26.2 &           39.8 &           38.4 &           53.9 & 20.6 \\
        OWL  \small (L/14)   &  \textbf{26.5} &           39.3 &  \textbf{44.0} &           65.1 & 31.2 \\
        OWLv2  \small (B/16) &           25.3 &           38.5 &           40.0 &           52.9 & 29.6 \\
        OWLv2  \small (L/14) &           25.4 &  \textbf{41.2} &           42.8 &           63.2 & 34.9 \\
        Detic        &           11.5 &           18.6 &           18.6 &  \textbf{69.7} & \textbf{39.9} \\
        ViLD         &           22.1 &           36.1 &           39.9 &           56.6 & 16.8 \\
        GDino        &           16.6 &           27.9 &           30.1 &           62.7 & 18.1 \\
        CORA         &           13.8 &           20.0 &           20.4 &           35.1 & 22.2 \\
        \bottomrule
        \end{tabularx}
        \caption{mAP on Difficulty-based benchmarks ($N=5$) and on rare categories of the standard LVIS benchmark (The mAP values for LVIS rare are taken from the original papers or GitHub repos).}
        \label{tab:map-benchmarks}
    \end{table}

    \begin{table}
        \centering
        \setlength{\tabcolsep}{2pt}
        \newcolumntype{C}{>{\centering\arraybackslash}X}
        \begin{tabularx}{\linewidth}{l*4C}
        \toprule
                            & \small   Color & \small Material & \small Pattern & \small Transp. \\
        \midrule
        OWL \small (B/16)   &           45.3 &           37.3 &           26.6 &  \textbf{34.1} \\
        OWL \small (L/14)   &           43.8 &  \textbf{44.9} &  \textbf{36.0} &           29.2 \\
        OWLv2 \small (B/16) &           45.1 &           33.5 &           19.2 &           28.5 \\
        OWLv2 \small (L/14) &  \textbf{53.3} &           36.9 &           23.3 &           12.2 \\
        Detic        &           21.5 &           38.8 &           30.1 &           28.0 \\
        ViLD         &           43.2 &           34.9 &           24.5 &           30.1 \\
        GDino        &           41.0 &           30.2 &           31.2 &           25.4 \\
        CORA         &           25.0 &           19.3 &           22.0 &           27.9 \\
        \bottomrule
        \end{tabularx}
        \caption{mAP on Attribute-based benchmarks ($N=2$).}
        \label{tab:map-attributes}
    \end{table}

    \begin{figure*}
        \centering
        \includegraphics[width=0.95\linewidth]{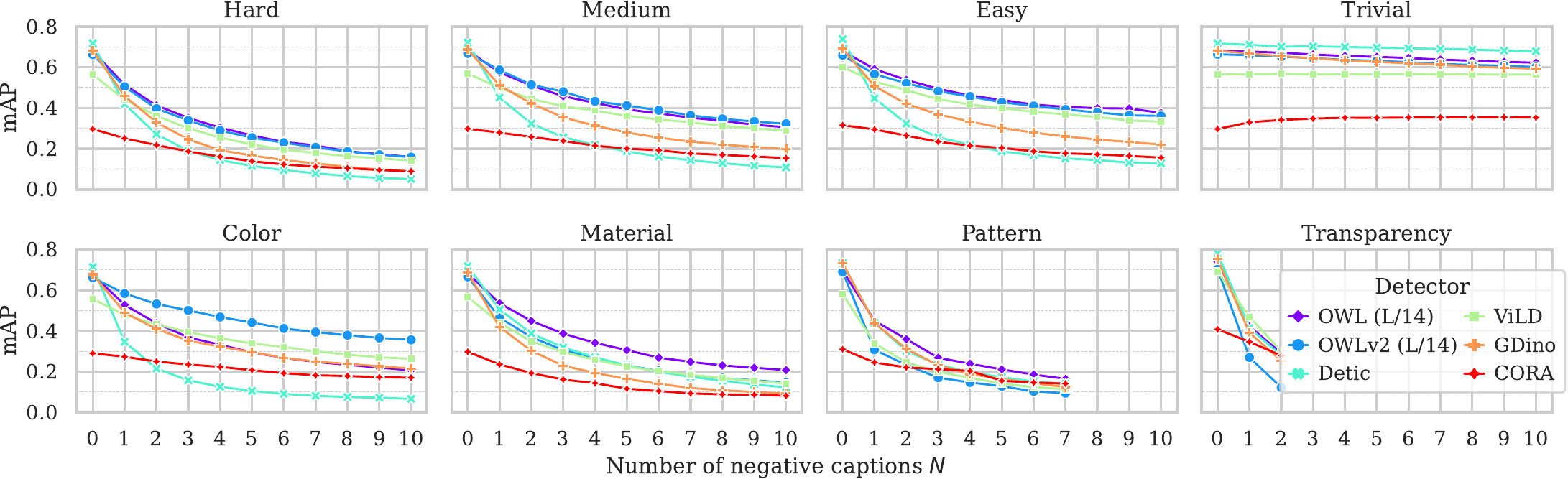}\\[2ex]
        \includegraphics[width=0.95\linewidth]{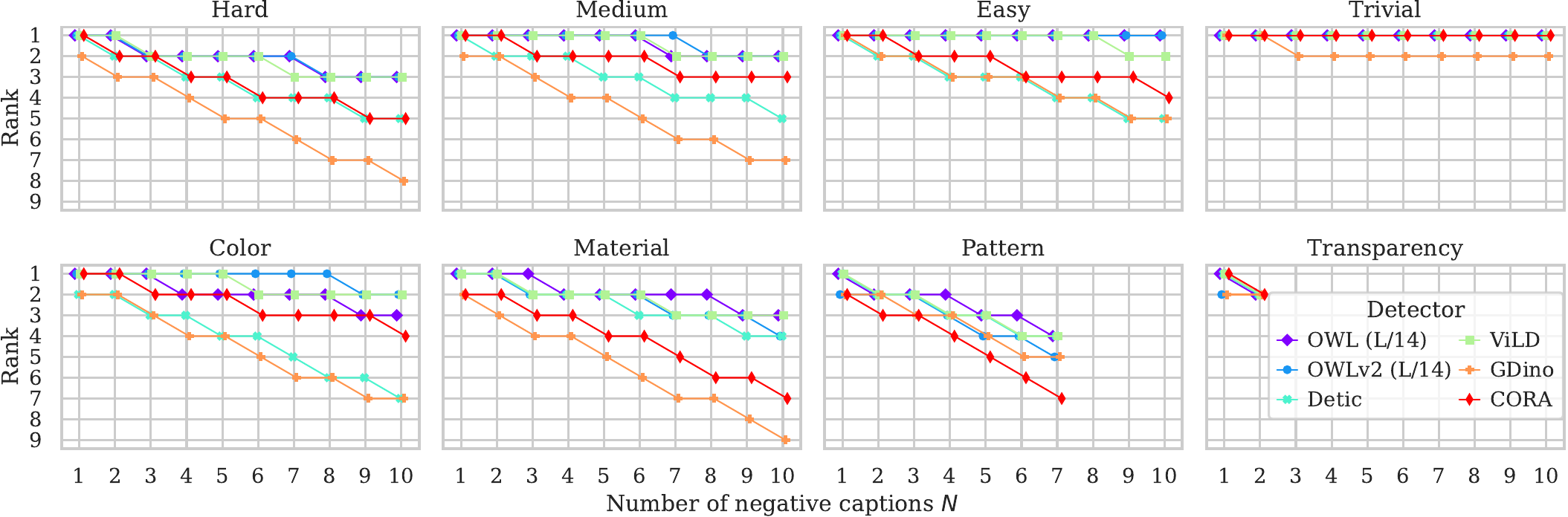}
        \caption{\textbf{Effect of the number of negative captions.} For each one of the eight proposed benchmarks, we report the mAP (rows 1-2) and the Rank (rows 3-4) varying the number $N$ of negative captions for the different probed detectors. Notice that for Pattern and Transparency, we have a limited number of possible negatives (7 and 2, respectively).}
        \label{fig:map-rank-vs-num-neg}
    \end{figure*}

\begin{figure*}[t]
        \includegraphics[width=\linewidth]{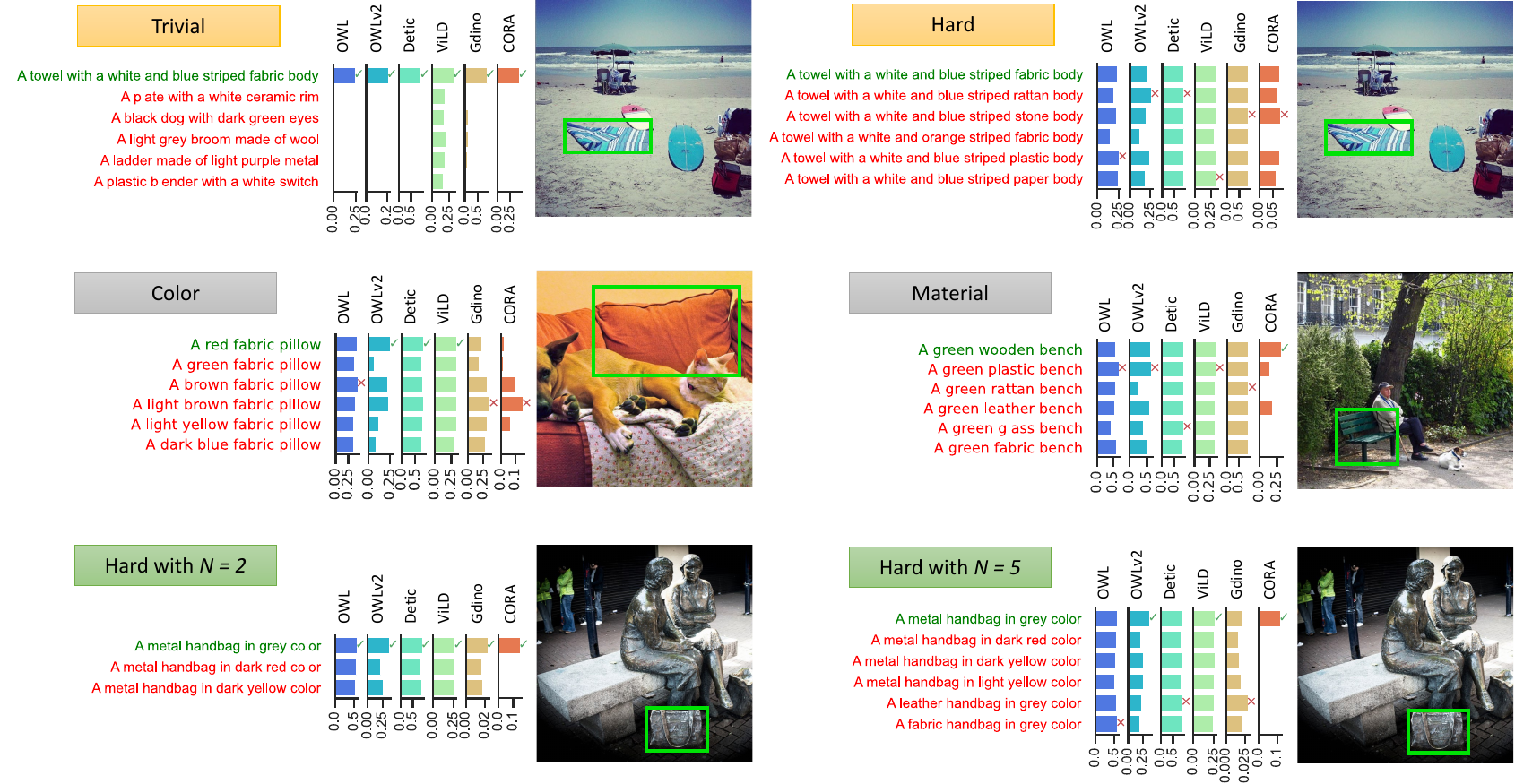}
        \caption{\textbf{Output scores from the probed detectors}. We report  examples of how detectors score vocabulary entries for a specific object. The first green caption is positive, while the other red ones are the negatives. \textbf{First row: Hard vs Trivial} -- we show the difference in score distributions when detectors are challenged with Hard (left) or Trivial (right). \textbf{Second row: Attributes} -- we show the behavior when changing specific attributes, like color (left) or material (right). \textbf{Third row: Varying the number of negatives} -- we show how increasing the number of negatives (left: $N=2$, right: $N=5$) strongly challenges the fine-grained discriminative abilities of many detectors.}
        \label{fig:predictions}
    \end{figure*}

    \paragraph{Performance vs Negative Difficulty.}
    \autoref{tab:map-benchmarks} shows the mAP of the tested state-of-the-art detectors on our difficulty-based benchmarks alongside reference results from existing benchmarks for standard open-vocabulary detection.
    We set the number of negative captions in the dynamic vocabularies $N=5$ for our benchmarks.
    In general, all detectors have acceptable performance in localizing and recognizing objects from detailed descriptions in the absence of confounding labels (Trivial), but the discriminative power in hard-negative settings drops drastically. For example, Detic is the best-performing detector on Trivial while the worst one in the Hard setting.
    Note also that performance in standard benchmarks does not positively correlate with performance in our hard-negative tasks. In this context, Detic is the detector performing better on LVIS while being the worst in the Hard scenario. 
    Generally, the best mAP in Hard is obtained by methods like OWL and ViLD that carefully embed image-language contrastively learned features into the detector heads. If properly managed inside the architecture, this information is crucial in giving the object detection heads discriminative abilities. Instead, Detic bases its strength on training with large image-level datasets, which add strong class-wise discriminative skills while largely sacrificing fine-grained attribute recognition abilities. This conclusion also seems motivated by the almost zero gain that OWLv2 -- trained in a self-supervised manner on web-scale data (10B images) -- has with respect to OWL. The training approach of OWLv2 possibly adds visual robustness to the already known classes but fails to inject proper discriminative attribute information despite the massive amount of data employed. Two qualitative examples in the first row of \autoref{fig:predictions} show how the scores entropy from the different detectors drastically increases when captions in the vocabulary start to contain very hard negatives.

    \paragraph{Performance vs Attribute Type.}
    In \autoref{tab:map-attributes}, we show the performance of each tested detector in discerning particular attribute types, by crafting hard negatives where only a particular attribute is changed.
    Note that we simplified the task for these results by adding only $N=2$ hard-negative captions to the vocabulary so that mAP values across different attributes are comparable\footnote{Different attributes imply different number of negatives, but N=2 is always guaranteed by our benchmark.}.
    We can notice that color is the easiest feature for most detectors, with OWLv2 reaching the best mAP. Color, in fact, is probably the attribute more present in web-scale image-text pairs (image and corresponding alt-text) used to train CLIP-like contrastive image-language models and to produce the N-grams vocabulary entries employed during the self-supervised training stage of OWLv2. On the other hand, it is possible that transparency and pattern properties are hardly found in image labels or alt-text captions collected in currently employed datasets. Despite good architectures and clever pre-training strategies, these attributes also undermine very recent open-vocabulary detectors like CORA. Two qualitative examples reported in the second row of \autoref{fig:predictions} show the sensitivity of different detectors to different attribute types.

    \paragraph{Performance vs Vocabulary Size.}
    \autoref{fig:map-rank-vs-num-neg} reports mAP and median rank for increasing the number of negative captions $N$ in the vocabulary up to 10 for each benchmark in the proposed suite.
    For Pattern and Transparency, the maximum number of negative captions $N$ is set to the number of available values for that attribute. %
    We note that, as expected, the detectors' performances degrade with increasing $N$ values, with different decreasing rates. Detic seems to have the steepest performance degradation, both in terms of mAP and Rank, as the number of negatives increases. Differently, despite an initial physiological decline, OWLv2 and ViLD seem to be the most robust on the Hard setting for the Color and Material attributes, despite remaining among the worst on Pattern and Transparency -- probably due to the absence of these attributes in their training and pre-training data.
    Notably, CORA seems challenged even for a small number of negatives, evidencing the trend that recent open-vocabulary detectors still suffer from major limitations when asked to distinguish fine-grained labels. Two qualitative examples reported in the third row of \autoref{fig:predictions} show how a higher number of negatives increases the scores distribution entropy, bringing higher misclassification rates. %

\section{Conclusions}

In this paper, we explored Fine-grained Open Vocabulary Detection (FG-OVD), by presenting a comprehensive evaluation protocol and benchmarks suite designed to scrutinize the fine-grained discriminative power of open-vocabulary detectors. The presented evaluation protocol, accompanied by meaningful metrics, challenges these models with rich captions encapsulating complex extrinsic characteristics. %

To implement the presented protocol, we prepared an ad-hoc benchmark, starting from structured object descriptions and employing an LLM to generate diverse and high-quality captions. By slightly changing one or more attributes in the generated caption, we produced a spectrum of difficulty levels, which enabled us to systematically analyze the weaknesses of recent open-vocabulary detectors across various analytical dimensions.
Our experiments revealed a notable gap in the detectors' ability to effectively capture and distinguish fine-grained object properties, with the most recent ones often performing the worst.

In the near future, we plan to fine-tune existing open-vocabulary detectors in a few-shot contrastive manner, tackling downstream tasks like episodic memory retrieval in egocentric datasets. Furthermore, we aim to exploit the proposed benchmarks to study latent information about fine-grained object attributes learned by diffusion models.

\paragraph{Acknowledgments}
This work was partially supported by the following projects:
SUN -- Social and hUman ceNtered XR (101092612),
NextGenerationEU (FAIR PE00000013,
ITSERR B53C22001770006,
MUCES B53D23026090001,
and
EKEEL P20227PEPK).

{
    \small
    \bibliographystyle{ieeenat_fullname}
    \bibliography{bibliography}
}

\clearpage


\maketitlesupplementary
\appendix

\section{Dataset details}
    The proposed benchmarks are based on PACO (Parts and Attributes of Common Objects), an attribute-based detection dataset.
    PACO covers 75 object categories, encompassing 456 object-part categories and 55 attributes across image and video datasets.
    The attributes used to describe objects and their parts are reported in the following table.\\[2ex]
    \vspace{1.5em}
    \begin{tabularx}{\linewidth}{rXXX}
    \toprule
    \textbf{Type} & \multicolumn{3}{l}{\textbf{Possible Values}} \\
    \midrule
    Colors & 
    black &
    light blue &
    blue \\ &
    dark blue &
    light brown &
    brown \\ & 
    dark brown &
    light green &
    green \\ &
    dark green &
    light grey &
    grey \\ &
    dark grey &
    light orange &
    orange \\ &
    dark orange &
    light pink &
    pink \\ &
    dark pink &
    light purple &
    purple \\ &
    dark purple &
    light red &
    red \\ &
    dark red &
    white &
    light yellow \\ &
    yellow &
    dark yellow \\ \midrule
    
    Materials &
    text &
    stone &
    wood \\ &
    rattan &
    fabric &
    crochet \\ &
    wool &
    leather &
    velvet \\ &
    metal &
    paper &
    plastic \\ &
    glass &
    ceramic \\ \midrule
    
    Patterns &
    plain &
    striped &
    dotted \\ &
    checkered &
    woven &
    studded \\ &
    perforated &
    floral &
    logo \\ \midrule
    
    Transp. & opaque & translucent & transparent \\
    \bottomrule
    \end{tabularx}\\[2ex]
    We simplified the structure of annotations in PACO to make it more straightforward for the LLM, aiming for more natural captions.
    Notably, we removed the \textit{plain} pattern and \textit{opaque} transparency from the object structure, as these are basic attributes found in almost every object without a specific pattern or transparency.
    Keeping them could lead to awkward sentences like \textit{A plain opaque black dog}, but we retained them for generation of negative captions.

    We also streamlined the structure by removing redundant attributes from object parts present in all components. Any missing attributes were added to the main object attributes to avoid overly complex sentences, and we also removed any part without attributes. For example, \textit{A car with a black hood, black roof, black fender, and black bumper} became \textit{A black car}.
    
    In the PACO dataset, only a subset of objects has attributes, while for others, the attributes are unknown. The dataset does not provide information about whether the attributes of one object also describe others in the same scene.
    This poses a potential problem, as a caption generated for one object may describe others not included in the ground truth of our benchmarks, confusing a potential positive as a negative and consequently poisoning the evaluation procedure.
    To address this issue, we initially propagated the generated caption for a given object indiscriminately to all the other objects having the same class. Then, objects inconsistent with the assigned captions were removed during manual revision.
    
\section{Captions generation} 
    \label{sec:captions_generations}
    
\begin{figure}[t]
        \centering
        \includegraphics[width=\linewidth]{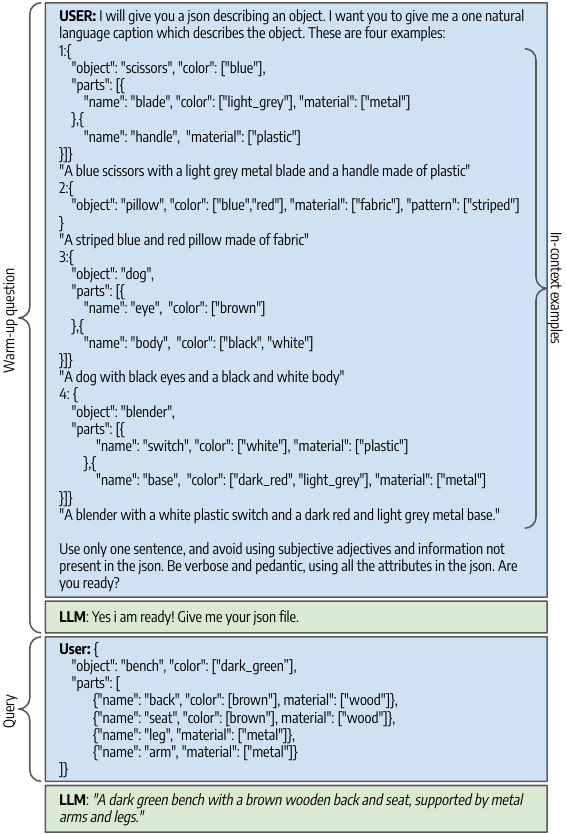}
        \caption{\textbf{Captions Generation.} 
        We query the LLM with a JSON description outlining the object's parts and attributes and four in-context samples to enhance caption quality and mitigate hallucination risks.
        We adopt a \textit{warm-up question} to separate the query from the in-context examples, as it avoids the LLM confusing the attributes of the examples with those of the query object. 
        }
        \label{fig:benchmark_generation}
    \end{figure}

\begin{figure}
        \centering
        \includegraphics[width=\linewidth]{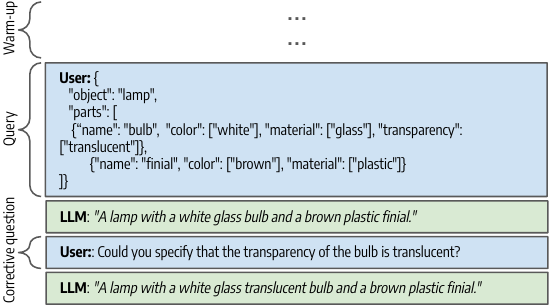}
        \caption{
            \textbf{Caption Correction.} If the initial caption did not meet our predefined criteria, we asked the LLM to generate an improved version, addressing the identified issues. The various follow-up questions for this iterative prompting process are outlined in \autoref{lst:iterative-prompting-requests}.}
        \label{fig:benchmark_regeneration}
    \end{figure}

    The generation process leverages prompt engineering and the in-context learning capabilities of LLMs \cite{brown2020language}.
    We present the model with pairs of object structures and corresponding natural language descriptions as illustrative examples.
    The model is then prompted to generate new captions based on queries describing the structural aspects of novel objects (see \autoref{fig:benchmark_generation}).
    In cases where the generated captions fail to meet predefined criteria, such as incomplete attribute utilization or excessive length, we initiate an automatic \textit{iterative prompting} process involving posing targeted follow-up questions to address empirically identified issues and refine the generated captions (see \autoref{fig:benchmark_regeneration}).
    We describe this approach in detail in \autoref{alg:create_caption}.
    We could apply this methodology indefinitely until all captions meet our criteria, but we limited to one iteration, as just a single one yielded sufficient data, removing the need for further iterations.
    
    Subsequent to caption generation, we meticulously conducted a manual revision process to uphold the benchmark's quality and accuracy.
    
    
    \begin{algorithm*}
    \caption{Caption Generation through Iterative Prompting}
    \label{alg:create_caption}
    \begin{algorithmic}[1]
    \Procedure{CreateCaption}{object, followup\_prompts, n\_iterations}
        \State $ \text{prompt} \gets \text{CreatePrompt(object)}$
        \Comment{Creates a prompt with 4 in-context examples + object structure}
        \State i $\gets$ 0
        \While{i $<$ n\_iterations}
            \State $ \text{caption} \gets \text{LLM(prompt)}$
            \State $ \text{identified\_problem} \gets \text{CheckIssues(caption, object)}$
            \Comment{Empirically check for issues in caption generation}
            \If{$ \text{identified\_problem is None}$}
                \State \textbf{return} $ \text{caption}$
                \Comment{The caption is correct}
            \EndIf
            \State $ \text{prompt} \gets \text{prompt} + \text{caption} + \text{followup\_prompts[identified\_problem]}$
            \State $ i \gets i + 1$
        \EndWhile
        \State \textbf{return} \text{None}
        \Comment{Caption incorrect after \textit{n\_iterations}}
    \EndProcedure
    \end{algorithmic}
\end{algorithm*}

\begin{figure*}
    \begin{lstlisting}[style=mystyle, caption={List of prompts employed in Iterative Prompting for correcting inaccurate captions. Each prompt is accompanied by a comment elucidating the condition that prompts the activation of that particular question.}, label={lst:iterative-prompting-requests}]
followup_prompts = {
    # caption with more than 60 words
    0: "Your (*@\textit{answer}@*) was too long. Create only one sentence for the object that describes what the object looks like considering its attributes",
    # ' is a ' inside the caption
    1: "Your answer is a definition of what the object is. Give me a caption that only describes the object and its attributes",
    # object not inserted
    2: "You did not specify that you are describing a {(*@\textit{object\_name}@*)}. Reformulate the caption with this addition", 
    # part not inserted
    3: "You did not specify that the {(*@\textit{object\_name}@*)} has a {(*@\textit{part\_name}@*)}. Reformulate the caption with this addition",
    # attribute not considered
    4: "Could you specify that the {(*@\textit{attribute\_type}@*)} of the {(*@\textit{object\_name}@*)} is {(*@\textit{attribute\_value}@*)}?",
    # ':' in the caption
    5: "Do not list the elements of the object. Summarize the description of the object in a natural language caption",
    # more than 2 '"'
    6: "You gave me more than one caption. Summarize them in only one caption",
    # a number in the caption
    7: "Your answer contains a number not present in the JSON. Create a new caption considering only the attributes I gave you and without adding information",
    # only one '"' in the caption
    8: "Answer is not complete. Write a complete caption",
    # found an illegal character
    9: "Illegal characters in the caption. Remove them",
    # 'or' in the caption
    10: "Ensure that the attributes are described using 'and' instead of 'or' to correctly represent all the specified attributes.",
    # 'single' in the caption
    11: "You used the word 'single'; reformulate the caption without it"
}
    \end{lstlisting}
\end{figure*}

\section{Model Architectural Details}
\begin{figure*}[htb]
    \centering
    \includegraphics[width=\linewidth]{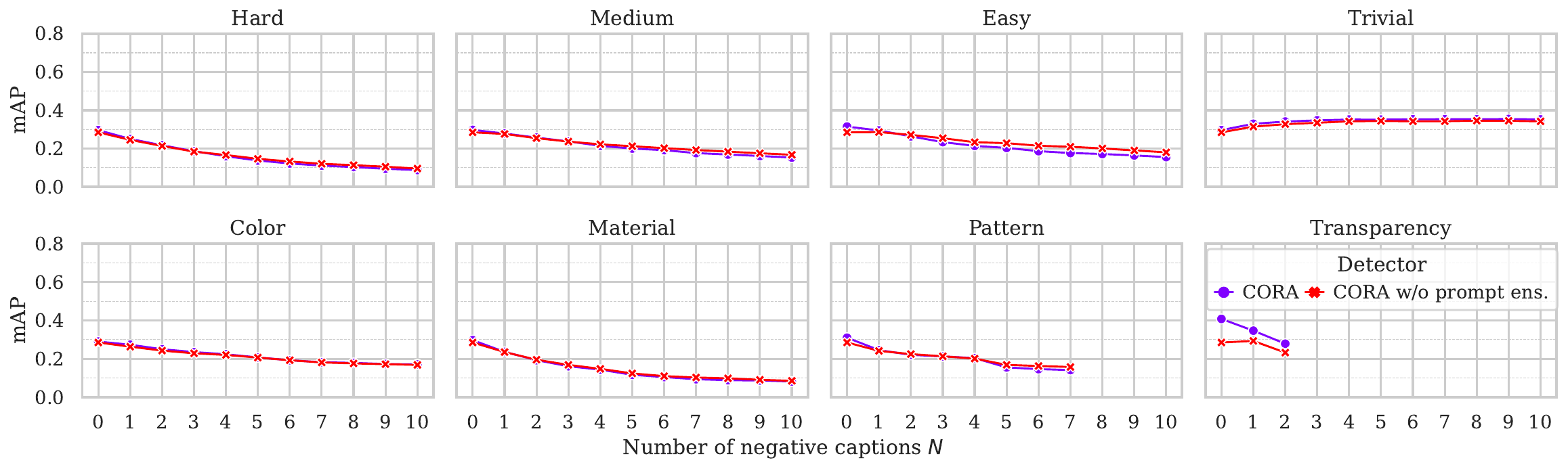}\\[2ex]
    \caption{\textbf{Effect of the number of the prompt ensemble on CORA}}
    \label{fig:map-vs-num-neg-CORA}
\end{figure*}

    In our experiments, we employed the following architectural configurations of detectors:

    \begin{compactitem}
    \item For OWL-based models (\textbf{OWL-ViT} and \textbf{OWLv2}), we evaluated configurations with ViT B/16 and ViT L/14 backbones.
    \item For \textbf{ViLD}, we utilized the configuration featuring Resnet-152 as a backbone, with a distillation weight set to 0.1.
    \item \textbf{Detic} was configured with the larger setup, employing Swin-B as the backbone and ImageNet-21K pre-training, specifically, the \texttt{Detic\_LCOCOI21k\_CLIP\_SwinB\_896b32\_4x\_ft4x \_max-size} configuration.
    \item \textbf{GroundingDino} was instantiated using the GroundingDINO-T configuration, employing Swin-T as the backbone.
    \item For \textbf{CORA}, we utilized the model with Resnet50x4 as the backbone. 
    \end{compactitem}
    As the captions in our benchmark are in natural language, all inferences were conducted without any pre-appended prompts. The sole exception is observed in \textit{CORA}, which employs an internal prompt ensemble, amalgamating 80 distinct prompts. In \autoref{fig:map-vs-num-neg-CORA}, we present the results of this model with prompt ensemble disabled, processing the input caption without any modifications. These results indicate that the performances are not significantly affected by the prompt ensemble.

\begin{figure*}[t]
        \centering
        \includegraphics[width=\linewidth]{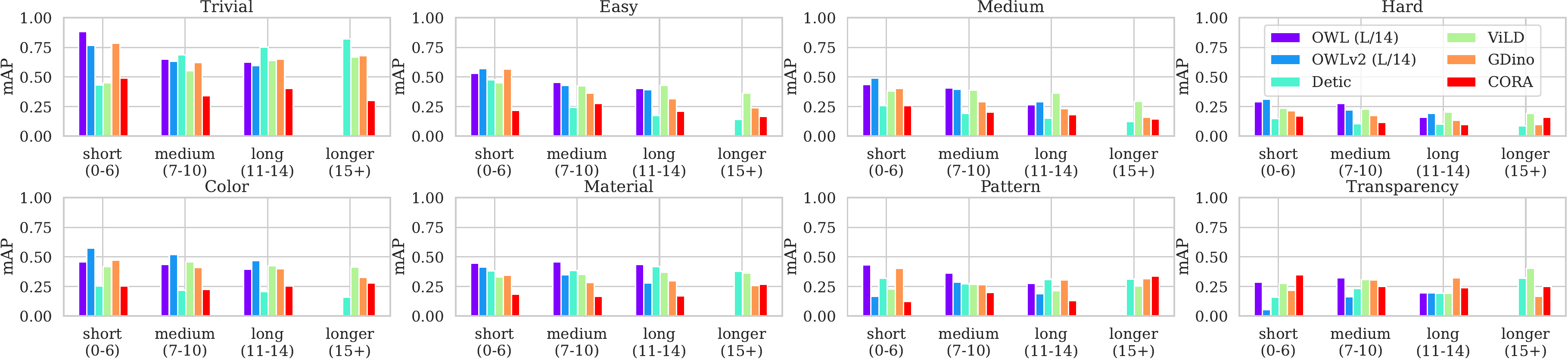}\\[2ex]
        \caption{\textbf{Effect of the caption length}: We illustrate the mAP of detectors across Difficulty-based (\textit{N = 5}) and Attribute-based (\textit{N = 2}) benchmarks, with varying caption lengths for objects. Captions are categorized into four groups based on their average word count inside the corresponding vocabulary: \textit{short} (6 or fewer words), \textit{medium} (7-10 words), \textit{long} (11-14 words), and \textit{longer} (15 or more words). OWL-based detectors are excluded from the \textit{longer} group due to their inability to process captions exceeding 16 tokens.}
        \label{fig:map-vs-length}
    \end{figure*}
    
\section{Handling Grounding Dino}
    GroundingDino diverges from other detectors in its approach, as it is mainly conceived for REC and cannot be fed directly with a dictionary of captions but only with a single caption. For this reason, we made an inference for each caption within the vocabulary. Each prediction is then represented as $d_i = (\mathbf{b}_i, h_i, t_i)$, with $\mathbf{b}_i$ being the bounding box coordinate and $h_i$ denoting the score value assigned to the caption $t_i$. This differs from other open-vocabulary object detectors, where each prediction incorporates a score array $\mathbf{s}_i$, and each element reflects the score associated with the corresponding caption. Importantly, this distinction does not impact the mAP of the detector, as it solely considers the predicted label for each corresponding predicted box. However, this difference affects the rank metric, as the score array $\mathbf{s}_i$ over the vocabulary entries is needed.
    
    For the rank calculation, where we need to rank all the possible vocabulary elements for each object, we considered all generated predictions $d_{j}$ for the ground truth object $o_i$, without employing the class-agnostic NMS typically applied in our evaluation protocol. Instead, we set to zero the confidence score of predictions not overlapping with the ground truth, i.e., $h_j \gets 0$ if $IoU(o_i, d_{j}) < 0.5$. 
    The score $\mathbf{s}_i$ on which the rank is calculated is derived by considering, for all elements in the vocabulary of $o_i$, the prediction score associated with the caption having the highest confidence. Subsequently, the median rank is computed following the same procedure employed for the other detectors.

\section{Additional Results: mAP Small, Medium and Large}
    The data in \autoref{tab:dimensional_map_n_attributes} and \autoref{tab:dimensional_map_types_attributes} offers a comprehensive view of mAP scores across Difficulty-based and Attribute-based benchmarks, segmented by object sizes. As expected, the task complexity increases with smaller-area objects, as finer details become more discernible with larger objects.
    
    \begin{table*}[htb]
    \centering
    \caption{Mean Average Precision (mAP) of detectors on the Hard, Medium, Easy and Trivial sets of negatives (N=5), segmented by object sizes (mAP$_\text{S}$, mAP$_\text{M}$, mAP$_\text{L}$)}
    \label{tab:dimensional_map_n_attributes}
    \newcolumntype{C}{>{\centering\arraybackslash}X}
    \begin{tabularx}{\linewidth}{l*{12}C}
    \toprule
    {} & \multicolumn{3}{c}{Hard} & \multicolumn{3}{c}{Medium} & \multicolumn{3}{c}{Easy} & \multicolumn{3}{c}{Trivial} \\
    \cmidrule(lr){2-4} \cmidrule(lr){5-7} \cmidrule(lr){8-10} \cmidrule(lr){11-13}
    {} &
    \small mAP$_\text{S}$ &
    \small mAP$_\text{M}$ &
    \small mAP$_\text{L}$ &
    \small mAP$_\text{S}$ &
    \small mAP$_\text{M}$ &
    \small mAP$_\text{L}$ &
    \small mAP$_\text{S}$ &
    \small mAP$_\text{M}$ &
    \small mAP$_\text{L}$ &
    \small mAP$_\text{S}$ &
    \small mAP$_\text{M}$ &
    \small mAP$_\text{L}$ \\
    \midrule
    OWL (B/16)   &          11.0 &          21.7 & \textbf{28.3} &          12.5 &          33.0 &          43.1 &          11.6 &          29.4 &          44.3 &          15.0 &          44.6 &          60.1 \\
    OWL (L/14)   &          13.8 &          22.7 &          27.0 &          18.3 &          34.4 &          39.9 &          19.3 &          39.3 &          42.9 &          33.0 &          56.0 &          67.9 \\
    OWLv2 (B/16) &          14.8 &          21.9 &          26.4 &          19.5 &          34.2 &          38.8 &          15.9 &          33.4 &          44.0 &          26.3 &          48.0 &          54.6 \\
    OWLv2 (L/14) &          14.0 &          21.9 &          26.2 & \textbf{24.2} &          34.6 & \textbf{43.9} &          13.8 &          34.3 & \textbf{48.6} &          32.0 &          55.8 &          65.8 \\
    Detic        &          10.5 &          11.5 &          12.2 &          11.6 &          20.2 &          18.4 &          19.9 &          19.3 &          19.1 & \textbf{39.4} & \textbf{71.1} & \textbf{75.0} \\
    ViLD         & \textbf{15.0} & \textbf{24.0} &          23.2 &          21.6 & \textbf{40.0} &          38.5 & \textbf{20.8} & \textbf{43.0} &          44.8 &          32.5 &          63.2 &          61.8 \\
    GDino        &           5.1 &          17.0 &          17.0 &           6.6 &          28.2 &          29.8 &           7.7 &          30.4 &          31.9 &          19.1 &          58.5 &          72.4 \\
    CORA         &           4.2 &          13.2 &          17.3 &           7.6 &          19.1 &          25.7 &           6.0 &          18.5 &          26.5 &          12.0 &          32.9 &          46.4 \\
    \bottomrule
    \end{tabularx}
    \end{table*}
    
    \begin{table*}[htb]
    \centering
    \caption{Mean Average Precision (mAP) of detectors on the Color, Material, Transparency and Pattern sets of negatives (N=2), segmented by object sizes (mAP$_\text{S}$, mAP$_\text{M}$, mAP$_\text{L}$)}
    \label{tab:dimensional_map_types_attributes}
    
    \newcolumntype{C}{>{\centering\arraybackslash}X}
    \begin{tabularx}{\linewidth}{l*{12}C}
    \toprule
    {} & \multicolumn{3}{c}{Color} & \multicolumn{3}{c}{Material} & \multicolumn{3}{c}{Transparency} & \multicolumn{3}{c}{Pattern} \\
    \cmidrule(lr){2-4} \cmidrule(lr){5-7} \cmidrule(lr){8-10} \cmidrule(lr){11-13}
    {} &
    \small mAP$_\text{S}$ &
    \small mAP$_\text{M}$ &
    \small mAP$_\text{L}$ &
    \small mAP$_\text{S}$ &
    \small mAP$_\text{M}$ &
    \small mAP$_\text{L}$ &
    \small mAP$_\text{S}$ &
    \small mAP$_\text{M}$ &
    \small mAP$_\text{L}$ &
    \small mAP$_\text{S}$ &
    \small mAP$_\text{M}$ &
    \small mAP$_\text{L}$ \\
    \midrule
    OWL (B/16)   &          15.6 &          39.1 &          48.2 &          11.2 &          29.7 &          42.5 &          17.0 & \textbf{31.3} &          37.5 &           2.7 &          24.8 &          28.3 \\
    OWL (L/14)   &          24.1 &          37.1 &          45.7 & \textbf{22.0} &          39.0 & \textbf{47.0} &          20.1 &          26.2 &          30.7 &          11.2 & \textbf{34.6} & \textbf{35.4} \\
    OWLv2 (B/16) &          24.5 &          42.1 &          44.6 &          14.9 &          27.1 &          37.9 & \textbf{20.2} &          26.6 &          29.7 &          18.0 &          19.8 &          20.5 \\
    OWLv2 (L/14) & \textbf{32.2} &          47.1 & \textbf{53.9} &          17.1 &          30.9 &          40.6 &           3.9 &           8.5 &          15.8 &          11.2 &          24.4 &          22.8 \\
    Detic        &          16.1 &          23.2 &          21.1 &          20.1 & \textbf{39.4} &          42.2 &           0.0 &          28.9 & \textbf{40.0} & \textbf{27.6} &          33.5 &          30.5 \\
    ViLD         &          26.3 & \textbf{49.6} &          44.7 &          16.7 &          37.6 &          39.9 &          13.9 &          30.9 &          34.0 &          13.0 &          28.9 &          24.0 \\
    GDino        &          12.1 &          39.4 &          45.9 &           7.0 &          28.0 &          34.4 &          19.7 &          22.8 &          27.6 &           7.6 &          26.9 &          34.7 \\
    CORA         &           9.4 &          24.8 &          32.5 &           7.7 &          17.0 &          26.8 &          10.8 &          25.8 &          34.9 &           4.2 &          21.9 &          28.2 \\
    \bottomrule
    \end{tabularx}
    \end{table*}

\section{Additional Results: Role of caption lengths}
    Our generated captions have a significant variability in terms of length, which also depends on the particular benchmark employed.
    Therefore, there exists the possibility that the results reported in the paper are correlated with the sentence length. For this reason, we report in \autoref{fig:map-vs-length} the mAP evaluated on subsets generated by grouping the sentences by length (intended as the average number of words in the annotation vocabulary).
    
    These results show that, on average, the difficulty of the task increases slightly as the caption length increases, as indicated by the weak negative correlation (Pearson coefficient) between length and mAP:

    \noindent
    {
    \small
    \vspace{2mm}
    \newcolumntype{C}{>{\centering\arraybackslash}X}
    \begin{tabularx}{\linewidth}{ccCCCC}
    OWL \tiny (L/14) & OWLv2 \tiny (L/14) & Detic & ViLD & GDino & CORA \\ \midrule
    -0.34 & -0.22 & -0.02 & 0.05 & -0.31 & -0.08
    \end{tabularx}
    }
    
    It is interesting to note that, while OWL-based detectors and GroundingDINO show a moderate effect of caption length on mAP, ViLD, Detic, and CORA show greater resilience to such variations. In general, we observe that the overall correlation remains quite bounded, meaning that caption length does not consistently affect the final results. 
\section{Additional Benchmark Samples}
    
    We show additional samples from our benchmarks in Figures
    \ref{fig:additional-samples:color} \textsf{(\capitalisewords{color})},
    \ref{fig:additional-samples:material} \textsf{(\capitalisewords{material})},
    \ref{fig:additional-samples:pattern} \textsf{(\capitalisewords{pattern})},
    \ref{fig:additional-samples:transparency} \textsf{(\capitalisewords{transparency})},
    \ref{fig:additional-samples:trivial} \textsf{(\capitalisewords{trivial})},
    \ref{fig:additional-samples:easy} \textsf{(\capitalisewords{easy})},
    \ref{fig:additional-samples:medium} \textsf{(\capitalisewords{medium})}, and
    \ref{fig:additional-samples:hard} \textsf{(\capitalisewords{hard})}.

\section{OWL Subset}
\begin{figure*}
        \centering
        \includegraphics[width=\linewidth]{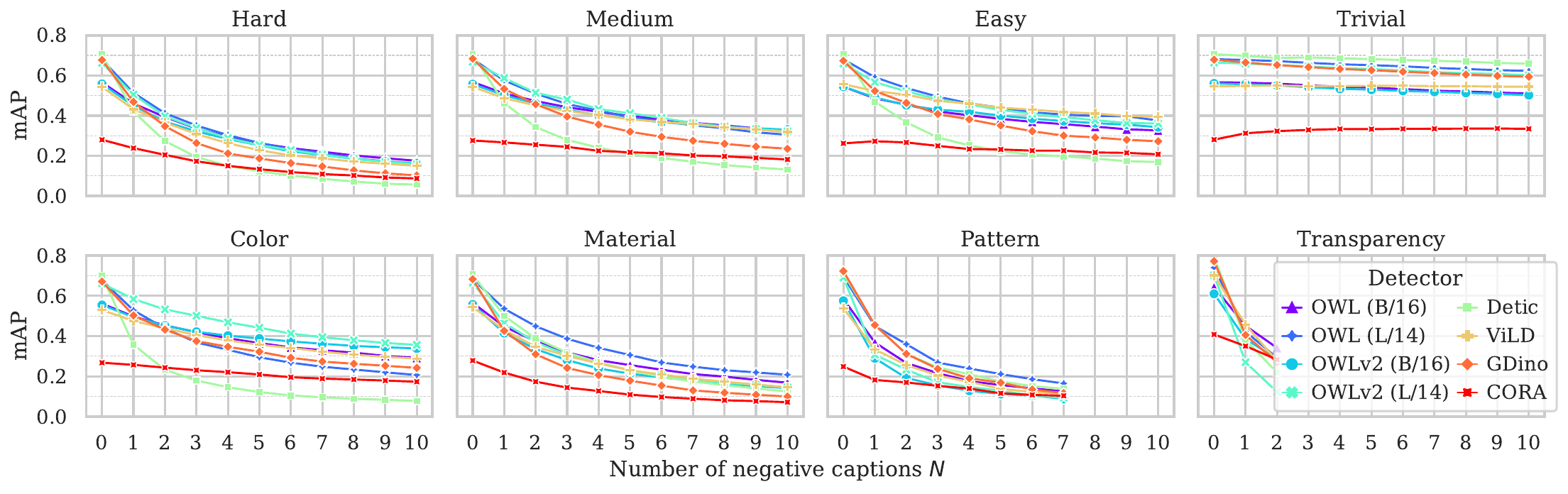}\\[2ex]
        \includegraphics[width=\linewidth]{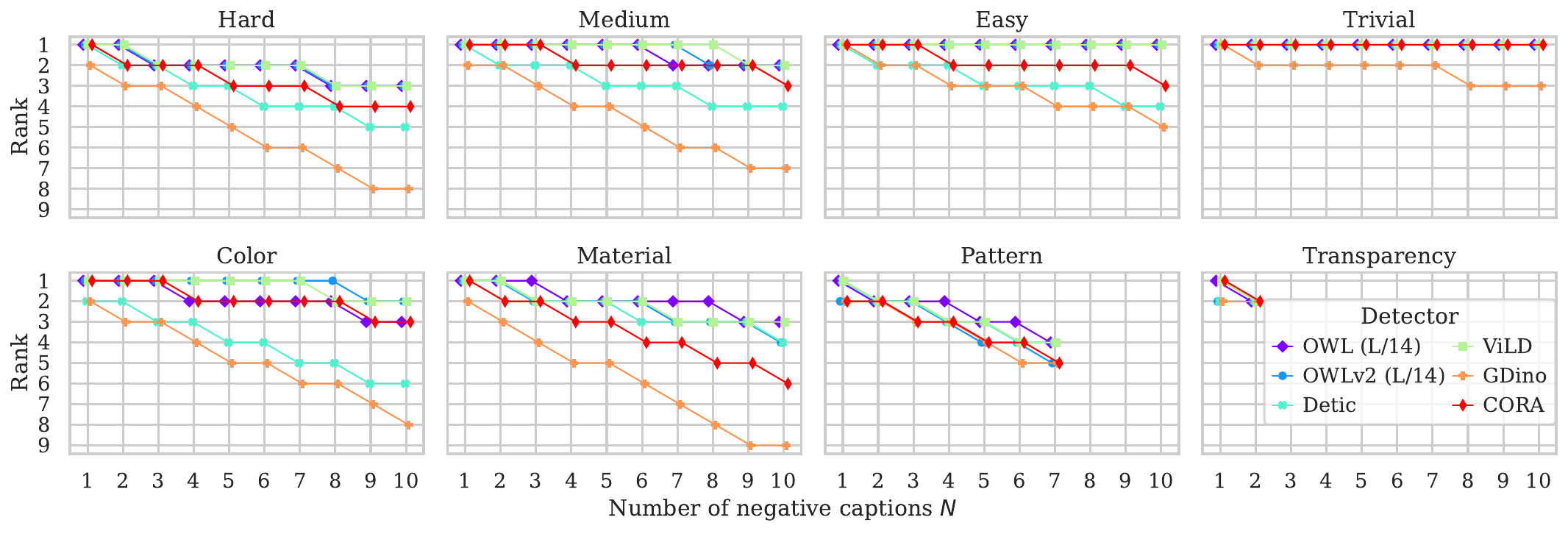}
        \caption{\textbf{Benchmark based on OWL-compatible captions:} Effect of the number of negative captions.}
        \label{fig:map-rank-vs-num-neg-OWL}
    \end{figure*}

    Since OWL processes sentences not exceeding 16 words, we tried to re-run the experiments over all the detectors with captions longer than 16 words removed.
    We present the updated statistics of the proposed benchmarks filtered using this constraint in \autoref{tab:owl_statistics}. 
    The corresponding results for all models on this subset are detailed in \autoref{tab:map-benchmarks-OWL}, \autoref{tab:map-attributes-OWL}, and \autoref{fig:map-rank-vs-num-neg-OWL}. Notably, due to the limited number of removed annotations, the results exhibit minimal deviation from those of the complete benchmarks, with each detector following a consistent trend. This suggests that the important information is likely placed early in the caption, and the last part in longer sentences can be ignored. 
    
    \begin{table*}
    \centering
    \caption{\textbf{Benchmark based on OWL-compatible captions:} Statistics of the benchmarks based on \textbf{OWL-subset} benchmark for each different negative set comprising the number of images (Imgs), the number of annotated objects (Objs), objects-to-image ratio (Objs/Img), positive captions, positive captions per image, negative captions per positive caption, and objects per positive caption.}
    \label{tab:owl_statistics}
    \begin{tabular}{llrrcrcrc}
    \toprule
    Name         & Negative Set Strategy                &  Imgs &    Objs &    Obj/Img & \cmark Caps & \cmark/Img & \xmark/\cmark &   Objs/\cmark \\
    \midrule
    Hard         & Random attribute subst. $(\times 1)$       &  1390 &  2903 &          2.1 &       1816 &            1.3 &          9.9 &           1.6 \\
    Normal       & Random attribute subst. $(\times 2)$       &  1187 &  2293 &          1.9 &       1483 &            1.2 &         10.0 &           1.5 \\
    Easy         & Random attribute subst. $(\times 3)$       &   417 &   657 &          1.6 &        445 &            1.1 &         10.0 &           1.5 \\
    Trivial      & Random captions  &  1389 &  2888 &          2.1 &       1810 &            1.3 &          10.0 &           1.6 \\
    \midrule
    Color        & Color attribute subst.              &  1269 &  2485 &          2.0 &       1595 &            1.3 &         10.0 &           1.6 \\
    Material     & Material attribute subst.           &  1277 &  2611 &          2.0 &       1639 &            1.3 &         10.0 &           1.6 \\
    Transparency & Transparency attribute subst.       &   177 &   323 &          1.8 &        180 &            1.0 &          2.0 &           1.8 \\
    Pattern      & Pattern attribute subst.            &   188 &   294 &          1.6 &        193 &            1.0 &          7.2 &           1.5 \\
    \bottomrule
    \end{tabular}
    \end{table*}

    \newcommand{\same}{} 
    \newcommand{\up}[1]{ \color{green}{\footnotesize (#1)}}
    \newcommand{\down}[1]{ \color{red}{\footnotesize (#1)}}
    
    \begin{table*}[t]
        \centering
        \begin{minipage}{.49\textwidth}
            \centering
            \setlength{\tabcolsep}{2pt}
            \newcolumntype{C}{>{\centering\arraybackslash}X}
            \begin{tabularx}{\linewidth}{l*4C}
            \toprule
                                & \small   Hard & \small Medium & \small Easy & \small Trivial \\
            \midrule
            OWL \small (B/16)   &           26.2\same &           39.8\same &           38.4\same &           53.9\same \\
            OWL \small (L/14)   &  \textbf{26.5}\same &           39.3\same &           44.0\same &           65.1\same \\
            OWLv2 \small (B/16) &           25.3\same &           38.5\same &           40.0\same &           52.9\same \\
            OWLv2 \small (L/14) &           25.4\same &  \textbf{41.2}\same &           42.8\same &           63.2\same \\ \midrule
            Detic        &           12.3\up{+0.8} &           20.9\up{+2.3} &           22.3\up{+3.7} &  \textbf{68.1}\down{-1.6} \\
            ViLD         &           22.8\up{+0.7} &           38.2\up{+2.1} &  44.0\up{+4.1} &           54.8\down{-1.8} \\
            GDino        &           18.7\up{+2.1} &           32.0\up{+4.1} &           35.1\up{+5.0} &           62.6\down{-0.1} \\
            CORA         &           13.4\down{-0.4} &           21.8\up{+1.8} &           23.2\up{+2.8} &           33.3\down{-1.8} \\
            \bottomrule
            \end{tabularx}
            \caption{\textbf{Benchmark based on OWL-compatible captions:} mAP on Difficulty-based benchmarks ($N=5$).}
            \label{tab:map-benchmarks-OWL}
        \end{minipage}
        \hfill
        \begin{minipage}{.49\textwidth}
            \centering
            \setlength{\tabcolsep}{2pt}
            \newcolumntype{C}{>{\centering\arraybackslash}X}
            \begin{tabularx}{\linewidth}{l*4C}
            \toprule
                                & \small   Color & \small Material & \small Pattern & \small Transp. \\
            \midrule
            OWL \small (B/16)   &           45.3\same &           37.3\same &           26.6\same &  \textbf{34.1}\same  \\
            OWL \small (L/14)   &           43.8\same &  \textbf{44.9}\same &  \textbf{36.0}\same &           29.2\same  \\
            OWLv2 \small (B/16) &           45.1\same &           33.5\same &           19.2\same &           28.5\same  \\
            OWLv2 \small (L/14) &  \textbf{53.3}\same &           36.9\same &           23.3\same &           12.2\same  \\ \midrule
            Detic        &           23.2\up{+1.7} &           38.8\same &           31.1\up{+1.0} &            21.8\down{-6.2}  \\
            ViLD         &           43.9\up{+0.7} &           34.7\down{-0.2} &           25.6\up{+1.1} &           27.6\down{-2.5}  \\
            GDino        &           43.2\up{+2.2} &           30.9\up{+0.7} &           31.1\down{-0.1} &           26.9\up{+1.5}  \\
            CORA         &           24.3\down{-0.7} &           17.3\down{-2.0} &           16.9\down{-5.1} &           28.4\up{+0.5}  \\
            \bottomrule
            \end{tabularx}
            \caption{\textbf{Benchmark based on OWL-compatible captions:} mAP on Attribute-based benchmarks ($N=2$).}
            \label{tab:map-attributes-OWL}
        \end{minipage}
    \end{table*}
    
\section{PACO-provided captions}
    The original PACO dataset offers a collection of 5,000 text entries corresponding to objects within the dataset. We run our evaluation protocol over an alternative benchmark built on these captions for completeness.
    
\begin{figure*}
        \centering
        \includegraphics[width=\linewidth]{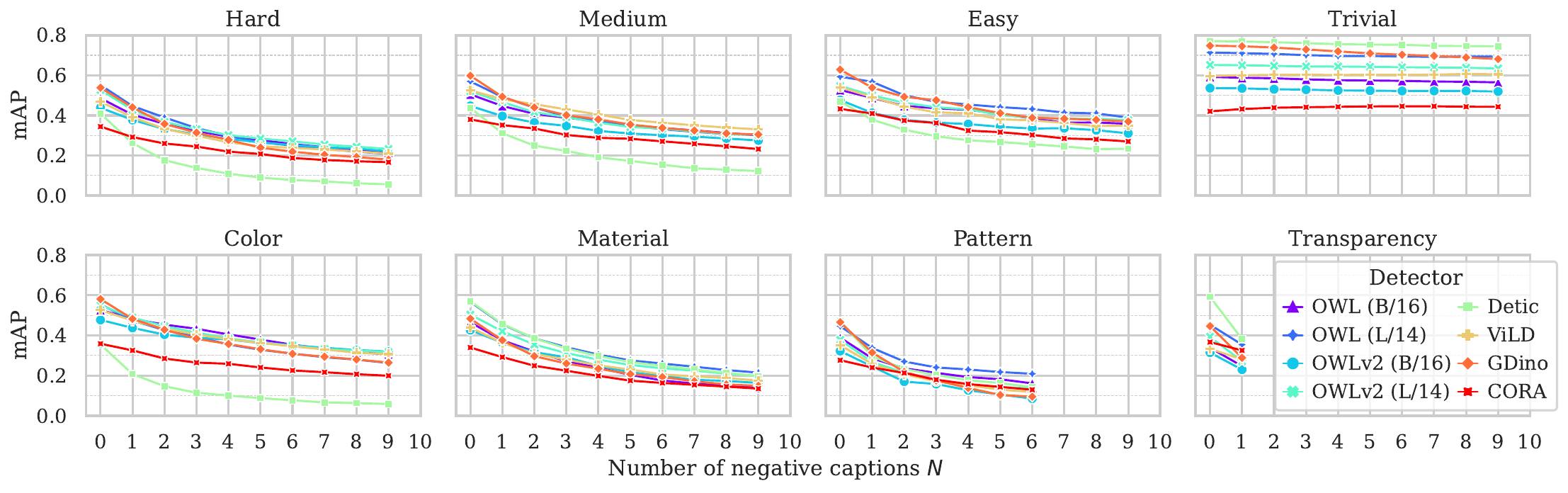}\\[2ex]
        \includegraphics[width=\linewidth]{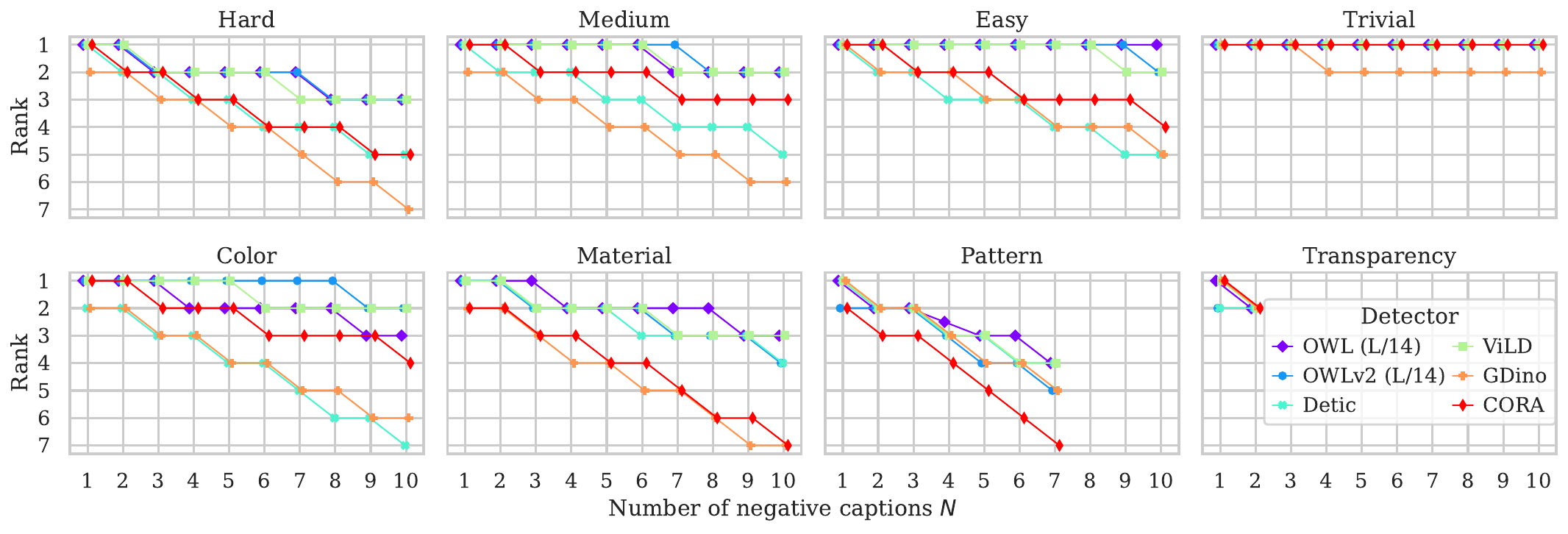}
        \caption{\textbf{Benchmark based on PACO-provided captions:} Effect of the number of negative captions.}
        \label{fig:map-rank-vs-num-neg-PACO}
    \end{figure*}

    To create this benchmark, we randomly selected one caption from the ones linked to each object, considering that each object could be associated with more than one caption.
    Notice that we needed to run a meticulous manual revision process like the one implemented for our official benchmark. In this process, the caption associated with an object was systematically allocated to all objects within the same image categorized under the identical class. 
    Subsequently, we manually assessed the objects, wherein objects discordant with the assigned caption were removed.
    We reported the statistics of the resulting benchmark in  \autoref{tab:paco_statistics}, and the results of the evaluated detectors in \autoref{tab:map_benchmarks-PACO}, \autoref{tab:map_attributes-PACO} and \autoref{fig:map-rank-vs-num-neg-PACO}. 
    
    However, in our main analysis, we still preferred to employ the captions generated using LLMs as described in \autoref{sec:captions_generations}. We followed such methodology for a series of reasons, explained in the following paragraphs.

\begin{figure*}
        \centering
        \includegraphics[width=\linewidth]{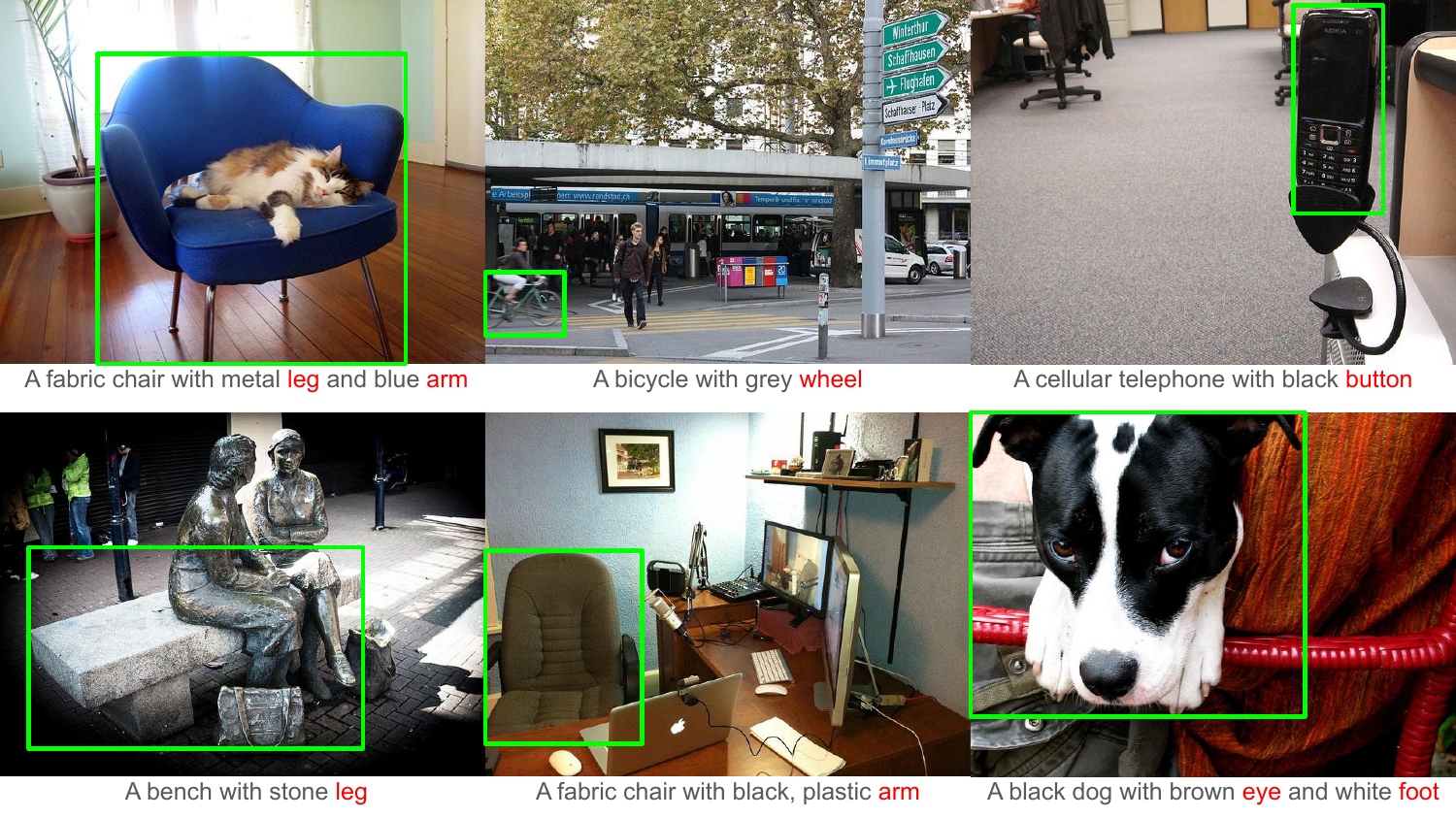}
        \caption{
        \textbf{Benchmark based on PACO-provided captions:} Samples where PACO captions do not include the plural form for parts. With its integrated common sense, a Large Language Model effectively addresses this by intuitively determining when pluralization is needed, resulting in sentences that feel more naturally structured.}
        \label{fig:paco-examples}
    \end{figure*}

    \paragraph{Limited Language Expressivity}
    PACO captions exhibit a tendency towards uniform syntactic structures, whereas we noticed that the utilization of an LLM introduces a welcomed variability. This variability facilitates the exploration of diverse natural language contexts, thereby enabling the evaluation of the detector in a broader array of scenarios. Furthermore, PACO captions occasionally manifest as linguistically unnatural --- i.e., parts are always singular even if there are multiple instances of the same part, as shown in \autoref{fig:paco-examples}. 

    \paragraph{Multiple Shorter Captions}
    Despite PACO's high number of captions, a single PACO object may be associated with multiple less detailed captions. For instance, a towel with attributes like black color and fabric material may be found encoded in three different captions, such as \textit{A black towel}, \textit{A fabric towel}, and \textit{A black fabric towel.} 
    This can be considered a limit in our scenario, where our evaluation protocol works by ingesting ad-hoc crafted negatives obtained by modifying a single attribute in a possibly long, detailed sentence.

    \paragraph{Limited Diversity}
    The quantity of object groups within benchmarks derived from PACO captions is notably limited. 
    The issue becomes apparent when examining the Transparency benchmark, which features a notably restricted number of object groups, as evidently shown in the Transparency row of \autoref{tab:paco_statistics}.
    This inherent scarcity is further exacerbated by OWL-based detectors being confined to a subset of each benchmark. Consequently, even a single error can induce substantial fluctuations in the measured mAP. \\

    \begin{table*}[t]
        \centering
        \begin{minipage}{.49\textwidth}
            \setlength{\tabcolsep}{2pt}
            \newcolumntype{C}{>{\centering\arraybackslash}X}
            \centering
            \begin{tabularx}{\linewidth}{l*4C}
            \toprule
                                & \small   Hard & \small Medium & \small Easy & \small Trivial \\
            \midrule
        OWL \small (B/16)   &           29.5 &           37.2 &           42.8 &           57.6 \\
        OWL \small (L/14)   &           29.9 &           37.2 &  \textbf{45.5} &           69.6 \\
        OWLv2 \small (B/16) &           28.3 &           32.3 &           35.7 &           52.5 \\
        OWLv2 \small (L/14) &  \textbf{30.2} &           36.4 &           43.0 &           64.4 \\
        Detic        &           11.0 &           19.1 &           27.6 &  \textbf{75.5} \\
        ViLD         &           26.6 &  \textbf{40.5} &           41.1 &           60.2 \\
        GDino        &           27.8 &           38.1 &           44.2 &           71.9 \\
        CORA         &           22.0 &           28.8 &           32.4 &           44.3 \\
            \bottomrule
            \end{tabularx}
            \caption{\textbf{Benchmark based on PACO-provided captions:} mAP on Difficulty-based benchmarks ($N=5$)}
            \label{tab:map_benchmarks-PACO}
        \end{minipage}
        \hfill
        \begin{minipage}{.49\textwidth}
            \centering
            \setlength{\tabcolsep}{2pt}
            \newcolumntype{C}{>{\centering\arraybackslash}X}
            \begin{tabularx}{\linewidth}{l*4C}
            \toprule
                                & \small   Color & \small Material & \small Pattern & \small Transp. \\
            \midrule
        OWL \small (B/16)   &           48.2 &           37.6 &           28.6 &           24.7 \\
        OWL \small (L/14)   &           47.5 &           45.5 &  \textbf{33.7} &           35.4 \\
        OWLv2 \small (B/16) &           43.8 &           36.4 &           24.5 &           22.9 \\
        OWLv2 \small (L/14) &  \textbf{48.9} &           42.1 &           26.8 &           27.1 \\
        Detic        &           20.7 &  \textbf{45.7} &           29.5 &  \textbf{38.2} \\
        ViLD         &           47.4 &           36.3 &           25.8 &           28.1 \\
        GDino        &           48.2 &           37.6 &           31.4 &           28.8 \\
        CORA         &           32.5 &           29.2 &           24.0 &           32.5 \\
            \bottomrule
            \end{tabularx}
            \caption{\textbf{Benchmark based on PACO-provided captions:} mAP on Attribute-based benchmarks ($N=2$)}
            \label{tab:map_attributes-PACO}
        \end{minipage}
    \end{table*}

    \begin{table*}
    \centering
    \caption{\textbf{Benchmark based on PACO-provided captions:} Statistics for each different negative set comprising the number of images (Imgs), the number of annotated objects (Objs), objects-to-image ratio (Objs/Img), positive captions, positive captions per image, negative captions per positive caption, and objects per positive caption.}
    \label{tab:paco_statistics}
    \begin{tabular}{llrrcrcrc}
    \toprule
    Name         & Negative Set Strategy                &  Imgs &    Objs &    Obj/Img & \cmark Caps & \cmark/Img & \xmark/\cmark &   Objs/\cmark \\
    \midrule
    Hard         & Random attribute subst. $(\times 1)$       &  1058 &  1326 &          1.3 &       1111 &            1.1 &          9.9 &           1.2 \\
    Normal       & Random attribute subst. $(\times 2)$       &   619 &   825 &          1.3 &        632 &            1.0 &         10.0 &           1.3 \\
    Easy         & Random attribute subst. $(\times 3)$       &   180 &   234 &          1.3 &        181 &            1.0 &         10.0 &           1.3 \\
    Trivial      & Random captions  &  1058 &  1326 &          1.3 &       1111 &            1.1 &          10.0 &           1.2 \\
    \midrule
    Color        & Color attribute subst.              &   901 &  1098 &          1.2 &        934 &            1.0 &         10.0 &           1.2 \\
    Material     & Material attribute subst.           &   464 &   615 &          1.3 &        476 &            1.0 &         10.0 &           1.3 \\
    Transparency & Transparency attribute subst.       &    90 &   113 &          1.3 &         90 &            1.0 &          2.1 &           1.3 \\
    Pattern      & Pattern attribute subst.            &   224 &   301 &          1.3 &        225 &            1.0 &          8.0 &           1.3 \\
    \bottomrule
    \end{tabular}
    \end{table*}

    \PredictionFigure{color}{attribute}
    \PredictionFigure{material}{attribute}
    \PredictionFigure{pattern}{attribute}
    \PredictionFigure{transparency}{attribute}
    \PredictionFigure{trivial}{difficulty level}
    \PredictionFigure{easy}{difficulty level}
    \PredictionFigure{medium}{difficulty level}
    \PredictionFigure{hard}{difficulty level}


\end{document}